%% file: mainv4.tex
\pdfoutput=1

\documentclass[11pt]{article}

\usepackage[final]{acl}

\usepackage{times}
\usepackage{latexsym}

\usepackage[T1]{fontenc}

\usepackage[utf8]{inputenc}

\usepackage{microtype}

\usepackage{inconsolata}

\usepackage{graphicx}
\usepackage{array}
\usepackage{multirow}
\usepackage{booktabs}
\usepackage{colortbl}
\usepackage{xcolor}
\usepackage{vcell}
\usepackage{arydshln}
\usepackage{amsmath}
\usepackage{amsfonts}
\usepackage{tabularx}
\usepackage{makecell}
\usepackage[normalem]{ulem}
\usepackage{CJKutf8}

\definecolor{lightblue}{RGB}{173, 216, 230}
\definecolor{lightred}{RGB}{255, 182, 193}

\definecolor{deepblue}{RGB}{50, 121, 168}
\definecolor{deepred}{RGB}{168, 70, 50}
\definecolor{deepgreen}{RGB}{59, 120, 66}

\definecolor{dpurple}{HTML}{1A0841}
\definecolor{dorange}{HTML}{D56516}

\definecolor{iterblue1}{RGB}{154, 201, 219}
\definecolor{iterblue2}{RGB}{61, 84, 136}

\title{%
  \includegraphics[width=0.44cm,height=0.44cm]{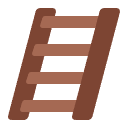}%
  \hspace{0.1cm}Ladder: A Model-Agnostic Framework Boosting LLM-based Machine Translation to the Next Level

}

\author{Zhaopeng Feng $^{1}$\thanks{Equally Contributed.}  \quad Ruizhe Chen $^{1}$$^{*}$  \quad   Yan Zhang$^{2}$\quad  Zijie Meng$^{1}$  \quad \bf Zuozhu Liu $^{1}$\thanks{Corresponding author.} \\
        $^{1}$ZJU-Angelalign R\&D Center for Intelligence Healthcare, Zhejiang University  \quad \\ 
        $^{2}$National University of Singapore\quad \\
        \texttt{\{zhaopeng.23, ruizhec.21, zijie.22 ,zuozhuliu\}@intl.zju.edu.cn} \\
        \texttt{eleyanz@nus.edu.sg} \\
}
\begin{document}
\maketitle
\begin{abstract}

General-purpose Large Language Models (LLMs) like GPT-4 have achieved remarkable advancements in machine translation (MT) by leveraging extensive web content. On the other hand, translation-specific LLMs are built by pre-training on domain-specific monolingual corpora and fine-tuning with human-annotated translation data. Despite the superior performance, these methods either demand an unprecedented scale of computing and data or substantial human editing and annotation efforts. In this paper, we develop \textbf{MT-Ladder}, a novel model-agnostic and cost-effective tool to refine the performance of general LLMs for MT. MT-Ladder is trained on pseudo-refinement triplets which can be easily obtained from existing LLMs without additional human cost. During training, we propose a hierarchical fine-tuning strategy with an easy-to-hard schema,  improving MT-Ladder's refining performance progressively. The trained MT-Ladder can be seamlessly integrated with any general-purpose LLMs to boost their translation performance. By utilizing Gemma-2B/7B as the backbone, MT-Ladder-2B can elevate raw translations to the level of top-tier open-source models (e.g., refining BigTranslate-13B with +6.91 BLEU and +3.52 COMET for XX→En), and MT-Ladder-7B can further enhance model performance to be on par with the state-of-the-art GPT-4. Extensive ablation and analysis corroborate the effectiveness of MT-Ladder in diverse settings. Our code is available at \href{https://github.com/fzp0424/MT-Ladder}{https://github.com/fzp0424/MT-Ladder}.

\end{abstract}

\begin{figure}[ht]
    \centering
    \vspace{-1mm}
\centerline{\includegraphics[width=\columnwidth]{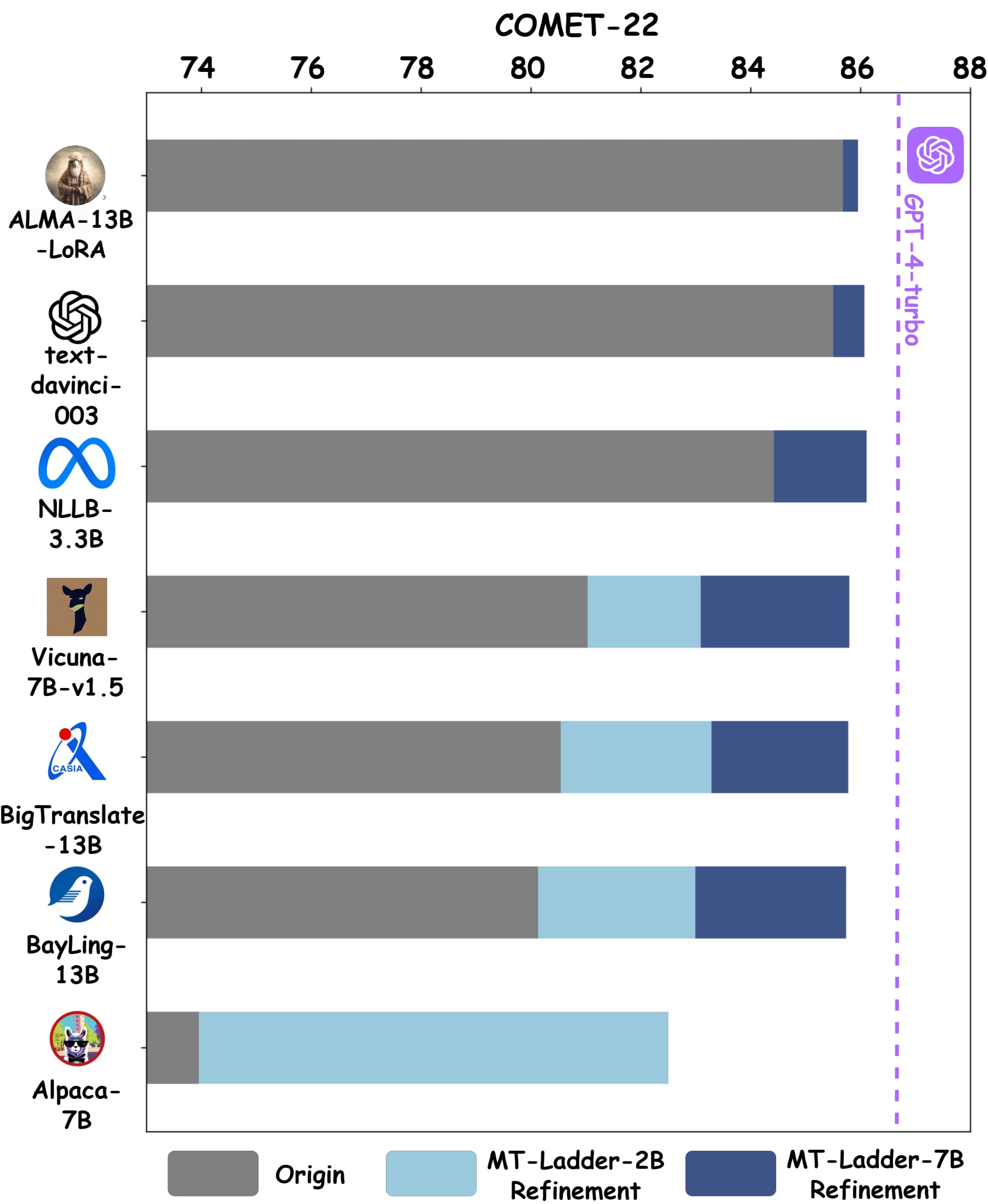}}
    \caption{ The average translation quality improvements across 8 translation directions on WMT22 test set (Zh$\leftrightarrow$En, De$\leftrightarrow$En, En$\leftrightarrow$Ru, En$\leftrightarrow$Cs) using MT-Ladder-2B or 7B. The metric
scores are calculated by COMET-22 (\textit{wmt22-comet-da})~\citep{rei2020comet}.} 
    \label{fig:intro}
    \vskip -0.2in
\end{figure}

\section{Introduction}

General-purpose Large Language Models (LLMs) like GPT-4~\citep{achiam2023gpt} have exhibited strong translation abilities~\citep{gpt-mt-2023, zhu2023multilingual, jiao2023chatgpt}, but achieving this performance requires enormous model scale, infrastructure, and deployment costs.  On the other hand, translation-specific LLMs like ALMA~\citep{xu2023paradigm} and Aya 23~\citep{aryabumi2024aya} have reached top-tier levels through continued pretraining on large monolingual corpora (e.g., 20B tokens from Common Crawl~\citep{OrtizSuarezSagotRomary2019}) and fine-tuning on high-quality translation data (e.g., 10.5M translation examples from Aya Dataset~\citep{singh2024aya}), which is also time-consuming and costly. These observations raise a question: \textit{can we enhance the MT performance of existing LLMs in a model-agnostic manner, achieving results comparable to translation-specific LLMs or even GPT-4, without incurring the significant costs associated with human annotations or extensive training?}

There are two potential approaches to  achieving this goal. The first is the prompt-based method, which involves developing effective prompting strategies to better stimulate LLMs' translation capabilities, such as using in-context translation examples, as outlined in works \citep{agrawal-etal-2023-context, pmlr-v202-garcia23a, peng-etal-2023-towards, chen2023iterative, feng2024improving}. However, \citet{zhang2023prompting} indicate that prompting methods overly rely on the language model, often under-translate the input and generate hallucinations. Additionally, 
\citet{moslem-etal-2023-adaptive} demonstrate that the same prompting strategy can lead to different performance across different models. Furthermore, most of these prompting strategies like agent debating or self-correction ~\citep{liang2023encouraging,feng2024improving} cannot be applied to some popular neural machine translation models like NLLB \citep{costa2022no}. These limitations make the learning-free method non-model-agnostic and unstable.



Another line of work employs learning-based paradigms by fine-tuning LLMs to adapt Quality Estimation (QE, \citealp{quality2010}) and Automatic Post-Editing (APE, \citealp{simard-etal-2007-rule}) tasks to refine raw translations. QE involves automatically predicting translation quality, typically using Multidimensional Quality Metrics (MQM) datasets~\citep{freitag2021experts}, where human experts annotate error spans and assign quality scores.  APE aims to address systematic errors of a black-box MT system and tailor the output to the lexicon and style required in a specific application domain. APE datasets are manually collected from real-world post-editing triplets like QT21~\citep{specia-etal-2017-translation}.  Built on these well-defined tasks and annotated datasets, prior works \citep{zeng2023tim,xu2023pinpoint,alves2024tower} have shown the promising utility and generalization of the learning-based method. \citet{xu2023pinpoint} trained PaLM2~\citep{anil2023palm} on MQM datasets to refine translations, and \citet{alves2024tower} trained TowerInstruct on 637k translation examples, integrating APE datasets, outperforming all open models and GPT-3.5-turbo on APE tasks. However, these works heavily rely on human-annotated evaluation data and lack extensive validation in model-agnostic and multilingual scenarios.  Additionally, the overall refinement in translation quality, particularly for translation-specific models, remains limited.



In this paper, we introduce \textbf{MT-Ladder}, a model-agnostic and cost-effective tool for multilingual translation refinement. Instead of directly fine-tuning a translation-target LLM, we train an LLM to refine translations using refinement datasets without human evaluation or post-edits, employing an instruction-following refinement task (Section~\ref{sec:ins-follow}). 
We notice that the \textit{reference} in existing parallel corpus can serve as a natural refined translation. By sampling a translation for the source sentence from an existing LLM as the \textit{intermediate translation}, we create a pseudo-refinement translation triplet [\textit{source, intermediate translation, reference}], allowing us to construct training data without extra labor costs. During training, we split the training triplets into three hierarchies (\textit{Easy}, \textit{Medium}, \textit{Hard}) based on their COMET~\citep{rei2020comet} scores and propose a hierarchical fine-tuning (HFT) strategy to improve MT-Ladder's refining performance step by step. Comprehensive experiments demonstrate that effectiveness of our MT-Ladder across various LLMs on multiple translation tasks. 

\begin{figure*}[ht]
    \centering
    \includegraphics[scale=0.21]{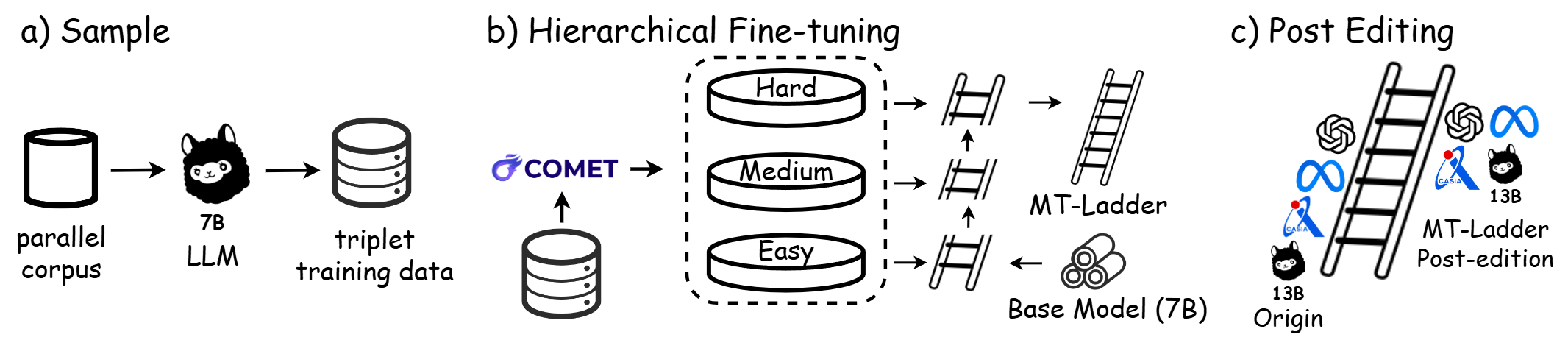}
    \caption{Obtain MT-Ladder in two steps: a) Sample from an LLM using the parallel corpus to create pseudo-refinement triplet training data. b) Use a hierarchical fine-tuning method with an easy-to-hard schema to tune the base model and obtain MT-Ladder. MT-Ladder can refine models with significantly higher parameter counts than the sampling LLM and base model. It can enhance original translations from various sources to the next level.} 
    \vspace{-3mm}
    \label{fig:method}
\end{figure*}

\section{MT-Ladder}



\subsection{Problem Formulation and Overview }
\label{sec:ins-follow}
\begin{figure}[ht]
    \centering
    \centerline{\includegraphics[width=\columnwidth]{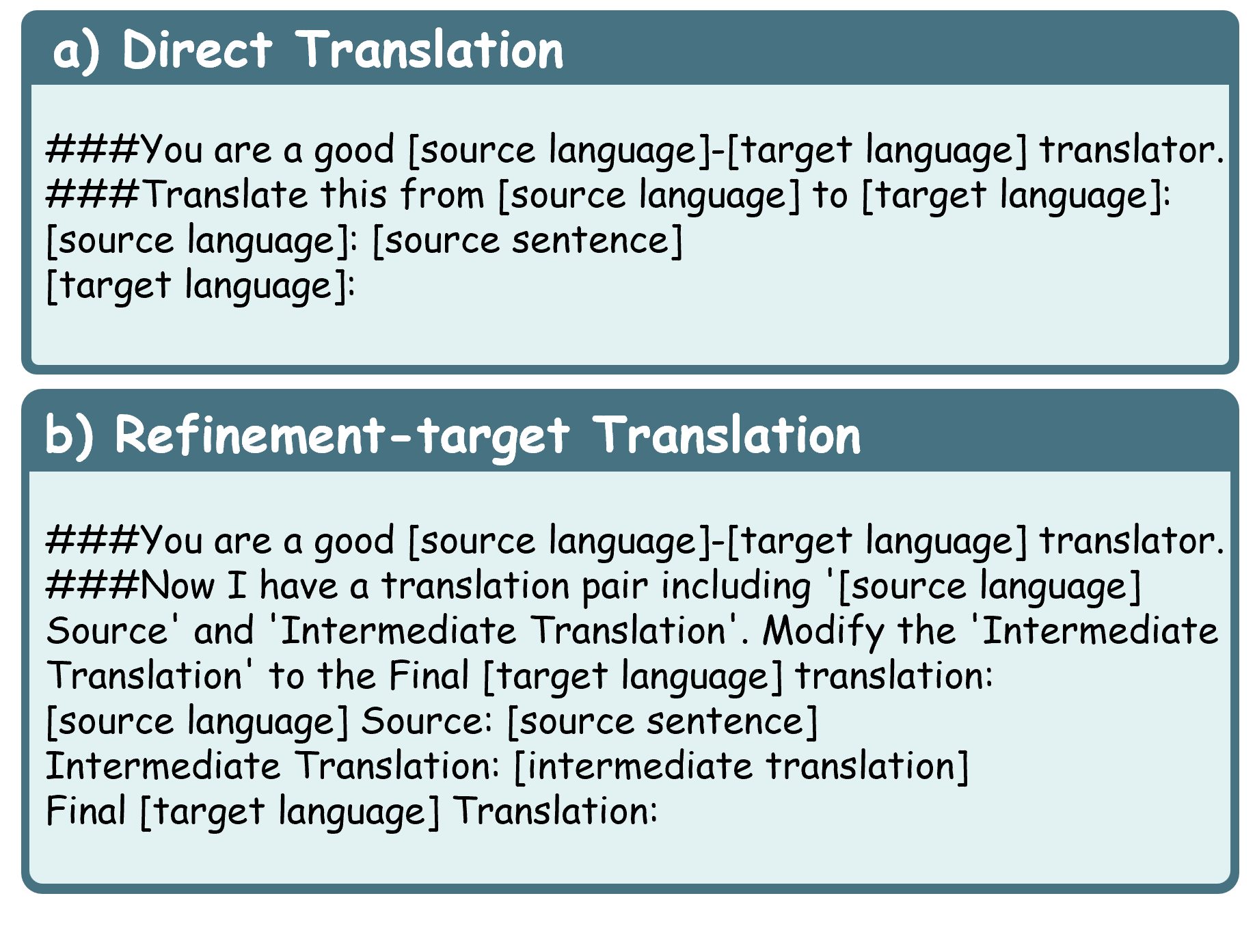}}
    \caption{Prompts used: [\textit{source language}] and [\textit{target language}] represent the full names of the languages. [\textit{source sentence}] is the sentence to be translated. [\textit{intermediate translation}] is the sampled translation. For Direction Translation, we follow \citet{xu2023paradigm}. }  
    \vspace{-3mm}
    \label{fig:prompt_all}
\end{figure}

Previous works \citep{zhang2023bayling, xu2023paradigm} adapt LLMs to translation tasks by fine-tuning on a parallel corpus [\textit{source}, \textit{reference}] using direct translation ($\mathcal{P}_{D}$) as shown in Figure~\ref{fig:prompt_all}. In contrast, we define our task as a refinement-target translation ($\mathcal{P}_{R}$) as shown in Figure~\ref{fig:prompt_all}, teaching the pre-trained base model to refine the existing translation of LLMs to the reference, rather than translating directly to the reference. Specifically, we introduce the concept of \textit{intermediate translation}, which denotes the translation sampled from existing LLMs. Then we add the intermediate translation to the pair [\textit{source}, \textit{reference}] to form a pseudo-refinement triplet [\textit{source}, \textit{intermediate translation}, \textit{reference}], taking the reference as the pseudo-refined translation.
The concept of translation refinement rather than direct translation is a key distinction of our work compared to previous translation-specific LLM approaches.

MT-Ladder models are created in two steps: 1) Sampling; and 2) Hierarchical Fine-tuning (HFT). First, given an existing LLM $\mathcal{M}_{S}$ and a parallel corpus $\mathcal{C}$, we use $\mathcal{M}_{S}$ to generate intermediate translations $i \sim {\mathcal{M}_{S}}(s,\mathcal{P}_{D})$ for each source sentence $s$ in the pair $(s,r) \in \mathcal{C}$, where $r$ is the reference. We then combine $i$ with $(s,r)$ to create pseudo-refinement triplets $(s,i,r)$, forming our training triplets $\mathcal{T}$. Second, we apply a hierarchical fine-tuning method with an easy-to-hard schema to fine-tune the base model on our instruction-following refinement task with triplet training data to obtain MT-Ladder $\mathcal{L}_{a}$. When applying $\mathcal{L}_{a}$ to refine the target LLM $\mathcal{M}_{T}$, $\mathcal{M}_{T}$ first generates the translation $i_{test} \sim {\mathcal{M}_{T}}(s_{test},\mathcal{P}_{D})$. $\mathcal{L}_{a}$ then refines $i_{test}$ into the final translation $y_{final} \sim {\mathcal{L}_{a}}(s_{test},i_{test},\mathcal{P}_{R})$. Figure~\ref{fig:method} shows the pipeline.

\subsection{Pseudo-refinement Triplet Construction}
Our pseudo-refinement triplet [\textit{source}, \textit{intermediate translation}, \textit{reference}] is similar in format to APE triplet [\textit{source}, \textit{translation with errors}, \textit{post-edits}]. However, the APE annotation procedure involves significant human costs for evaluation, error marking, and post-editing, focusing on word- or phrase-level corrections rather than overall translation quality improvement~\citep{specia-etal-2017-translation}. In contrast, our work uses reference $r$ as the supervised label, focusing on overall quality. Given the sampling LLM $\mathcal{M}_{S}$ with parameters $\theta_{S}$, parallel corpus $\mathcal{C}$ and prompt $\mathcal{P}_{D}$,  the intermediate translation $i$ for each pair $(s,r) \in \mathcal{C}$ can be generated auto-regressively as $i_{t} \sim p_{\theta_{S}} (i_{t} \mid s, \mathcal{P}_{D}, i_{<t})$. Naturally, the quality of $i$ is inferior to $r$, so we treat $r$ as the refined translation and construct our pseudo-refinement triplet training data $(s,i,r) \in \mathcal{T}$ without additional human costs.

\input{table/xxen_all}
\input{table/enxx_all}

\subsection{Hierarchical Fine-tuning}
\label{sec:hierarchical optimization}

Before fine-tuning, we use COMET \citep{rei2020comet} to categorize the pseudo-refinement triplet training data $\mathcal{T}$ into three levels: \textit{Easy}, \textit{Medium}, and \textit{Hard} and propose a hierarchical fine-tuning (HFT) strategy to achieve better refinement performance by learning from \textit{Easy} to \textit{Hard} examples.  \textit{Easy} translations differ significantly from the reference, offering the most room for refinement. \textit{Hard} translations are nearly perfect, with minimal differences, making them the hardest to refine. \textit{Medium} translations fall between these two poles. Translations with COMET scores below $\mu$ are classified as \textit{Easy}, scores between $\mu$ and $\nu$ as \textit{Medium}, and scores above $\nu$ as \textit{Hard}. We set thresholds $\mu$ and $\nu$ to 0.75 and 0.85, respectively, and analyze the effects of HFT and its robustness against these thresholds in Section \ref{sec:hierarchy_ft_effect}.

We fine-tune the pre-trained base model using instruction tuning (IT), aiming to obtain the model ${\mathcal{L}_{a}(\theta)}$ on pseudo-refinement triplet training data $\mathcal{T}=\{s^{(k)}, i^{(k)}, r^{(k)}\}_{k=1}^{N}$ by minimizing the following objective:
\begin{equation}
    \resizebox{0.8\columnwidth}{!}{$
    \mathcal{L}(\boldsymbol{\theta}; \mathcal{T}) = -\mathbb{E}_{(\boldsymbol{s}, \boldsymbol{i}, \boldsymbol{r}) \sim \mathcal{T}} \left[\log {\mathcal{L}_{a}}(\boldsymbol{r} \mid \boldsymbol{s},\boldsymbol{i}, \mathcal{P}_{R}; \theta) \right]
    $}
\end{equation}
We start with \textit{Easy} examples to help the base model capture detectable differences, then progressively fine-tune with the next level of examples, building on the previous stage.

\subsection{Translation Refinement}
When using MT-Ladder $\mathcal{L}_{a}$ with parameters $\theta_{\mathcal{L}_{a}}$ for refinement, given any target LLM $\mathcal{M}_{T}$ capable of translation, we first utilize $\mathcal{M}_{T}$ to generate the intermediate translation $i_{test} \sim {\mathcal{M}_{T}}(s_{test}, \mathcal{P}_{D})$. MT-Ladder then refines $i_{test}$ into the final translation $y_{final}$ in an auto-regressive manner: ${y_{final}}_{t} \sim p_{\theta_{\mathcal{L}_{a}}} ({y_{final}}_{t} \mid s_{test}, i_{test}, \mathcal{P}_{R}, {y_{final}}_{<t})$. 


\section{Experiments}

\subsection{Experimental Setup}

{\textbf{Datasets.}} For training, we choose Vicuna-7B-v1.5 \citep{vicuna2023} as the sampling model. Vicuna-7B-v1.5, fine-tuned from LLaMA2 \citep{touvron2023LLaMA}, possesses a certain level of translation ability (see Tables~\ref{tab:xxen} and~\ref{tab:enxx}). For parallel corpus, we collect test datasets from WMT'17 to WMT'20, along with Flores-200 \citep{costa2022no}, covering 8 translation directions (En $\Leftrightarrow$ XX) and 5 languages: English (En), German (De), Czech (Cs), Chinese (Zh), and Russian (Ru). The trained MT-Ladder is evaluated on the same translation directions using data from WMT22~\footnote{\href{https://github.com/wmt-conference}{https://github.com/wmt-conference}}. Detailed statistics are in Table~\ref{tab:data}.


\input{table/refine_cmp_v2}

We evaluate MT-Ladder under two scenarios. 1) We examine the effectiveness of MT-Ladder to refine both translation-specific LLMs, such as BigTranslate \citep{yang-etal-2023-BigTranslate}, BayLing \citep{zhang2023bayling}, NLLB \citep{costa2022no}, ALMA \citep{xu2023paradigm}, and general LLMs, such as Alpaca \citep{alpaca}, Vicuna \citep{vicuna2023}, GPT-3.5-text-davinci-003~\footnote{GPT-3.5 results are sourced from \citet{xu2023paradigm}.} \citep{ouyang2022training}, GPT-4~\footnote{GPT-4 results are sourced from \citet{xu2024contrastive}.} \citep{achiam2023gpt}. 2) We compare MT-Ladder to SoTA translation refinement or APE methods and models, i.e., Contrast Translation (CT) and Rephrase (Re) prompting strategies \citep{chen2023iterative}, LLMRefine \citep{xu2023pinpoint}, TowerInstruct \citep{alves2024tower} and API-based model GPT-4o mini. Details are in Appendix~\ref{app:Basemodel}.


\begin{figure*}[ht]
    \centering
    \includegraphics[scale=0.33]{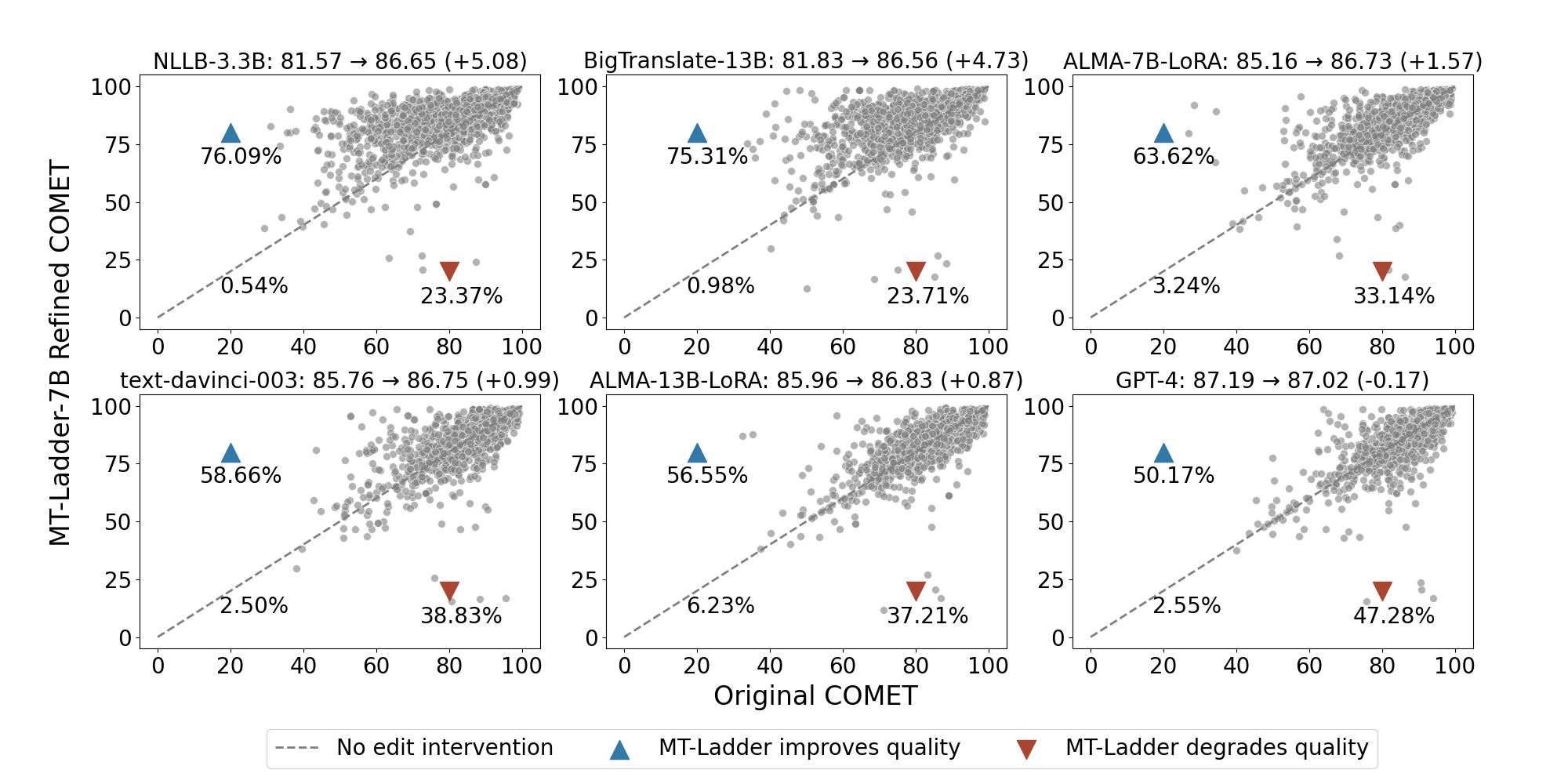}
    \caption{Comparison of original translation quality (x-axis) with MT-Ladder-7B refined quality (y-axis). Each dot is a WMT22 En-Zh translation. The percentages represent the proportion of each part, attached next to the markers.} 
    \vspace{-3mm}
    \label{fig:trend}
\end{figure*}

\begin{figure}[!ht]
    \centering

\centerline{\includegraphics[width=\columnwidth]{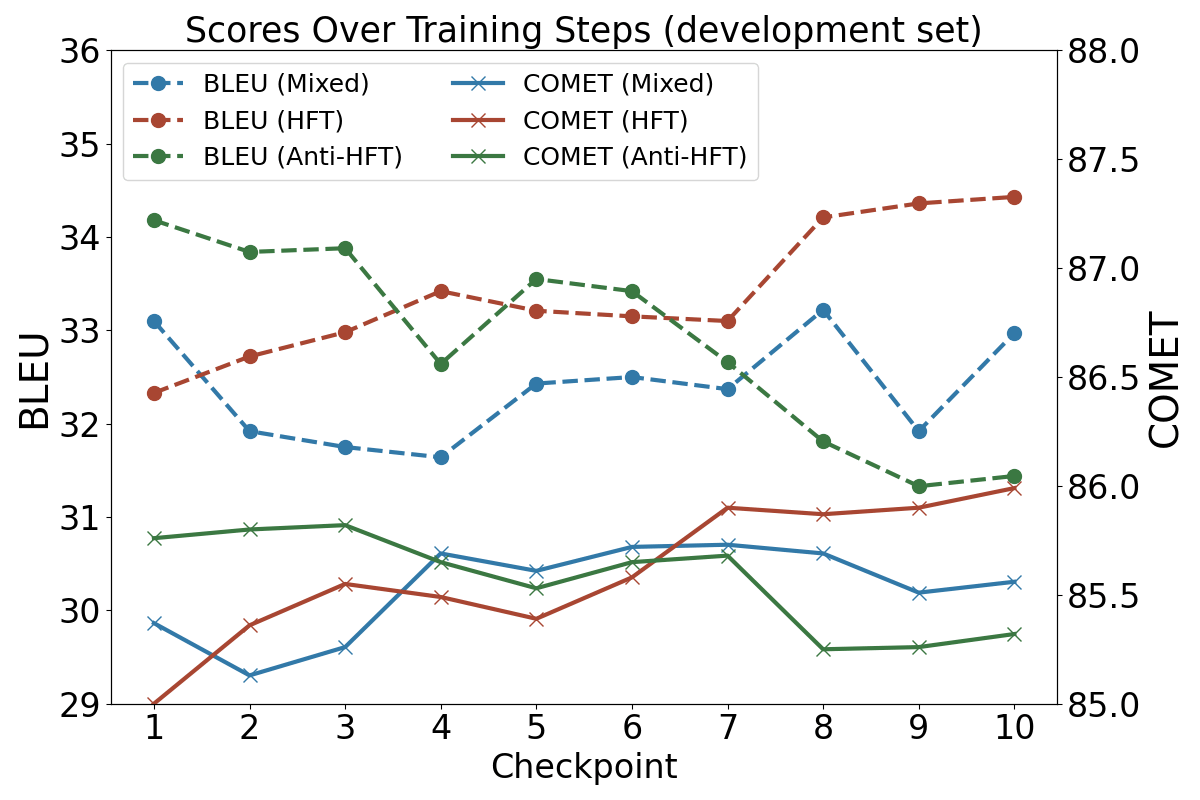}}
    \caption{Trends in BLEU and COMET during training. \textcolor{deepred}{HFT} represents our hierarchical fine-tuning from \textit{Easy} to \textit{Hard} examples, while \textcolor{deepblue}{Mixed} denotes using mixed data shuffling without hierarchical fine-tuning. \textcolor{deepgreen}{Anti-HFT} refers to reversing the HFT process.} 
    \label{fig:hierarchy}
    \vskip -0.2in
\end{figure}


\noindent {\textbf{Metrics.}} Following \citet{xu2023paradigm} and \citet{alves2024tower}, we use the lexical metric BLEU \citep{post-2018-call} and the reference-based metric COMET-22 \citep{rei2020comet} as the main metrics to evaluate the translation quality.  We further employ the reference-free QE model COMETKiwi \citep{rei2022cometkiwi} to evaluate the overall translation quality.

\noindent {\textbf{Backbones.}} MT-Ladder uses Gemma-2B and Gemma-7B\footnote{They utilize a vocabulary size of 256k tokens, ensuring effective applicability in multilingual scenarios.} as the backbones, which are further fine-tuned using LoRA \citep{hu2021lora} with a rank of 16. We update 0.9\% of the parameters for the 2B model and 0.6\% for the 7B model.\footnote{The training details are presented in Appendix~\ref{app:details}.} 



\subsection{Main Results}

\noindent {\textbf{Refinement Performance over LLMs.}}
Table \ref{tab:xxen} and \ref{tab:enxx} show that MT-Ladder can significantly improve the overall translation quality for all 8 translation directions across most translation-specific and general-purpose LLMs. Specifically, MT-Ladder-2B improves Alpaca-7B by $+12.07$ BLEU and $+13.25$ COMET for En$\rightarrow$XX on average, and refines BigTranslate-13B by $+6.91$ BLEU and $+3.52$ COMET for XX$\rightarrow$En. As for MT-Ladder-7B, it shows improvement over all open-source models on average. Notably, it even enhances 7 out of 8 translations for GPT-3.5-text-davinci-003 and improves $+1.05$ BLEU score for GPT-4 on average. We also find that while MT-Ladder-2B shows inferior performance on the strong NLLB-3.3B, our MT-Ladder-7B exhibits significant translation refinements on average. This aligns with our intuitions that different base models might exhibit varying levels of refinement performance across different LLMs, see detailed analysis in Figure~\ref{fig:trend}.

\noindent {\textbf{Comparison with SoTA Baselines.}} We compare MT-Ladder with SoTA baselines on four translation directions from WMT22, as reported in Table~\ref{tab:refine_cmp_v2}. We report the performance of LLMRefine on Palm2~\citep{xu2023pinpoint} as it is not available for refining BigTranslate. We can notice that MT-Ladder-7B significantly outperforms all open-source baselines and can even match the performance of GPT-4o mini. MT-Ladder-2B exhibits performance on par with the TowerInstruct-13B, which is superior than GPT-3.5-turbo~\citep{alves2024tower}. The results also highlight the instability of prompt-based methods. In contrast, MT-Ladder consistently demonstrates its lightweight and superior performance.

\begin{figure}[!ht]
\vskip 0.1in
    \centering
    \vspace{-2mm}
\centerline{\includegraphics[width=0.95\columnwidth]{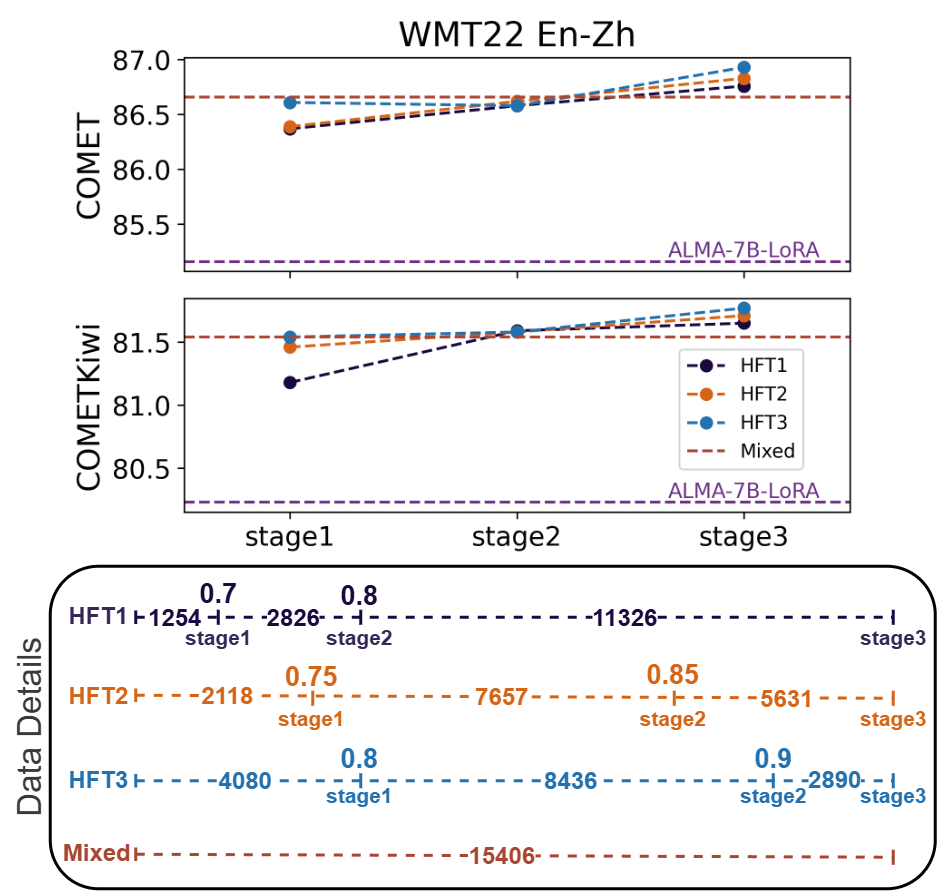}}
    \caption{Robustness against threshold $\mu$ and $\nu$. \textcolor{dpurple}{HFT1}: ($\mu$,$\nu$) = (0.7, 0.8), \textcolor{dorange}{HFT2}: ($\mu$,$\nu$) = (0.75, 0.85), and \textcolor{deepblue}{HFT3}: ($\mu$,$\nu$) = (0.8, 0.9). \textcolor{deepred}{Mixed} denotes mixed training. ALMA-7B-LoRA is the model to refine. }
    \label{fig:score_stages}
    \vskip -0.2in
\end{figure}

\begin{figure*}[ht]
    \centering
    \includegraphics[scale=0.28]{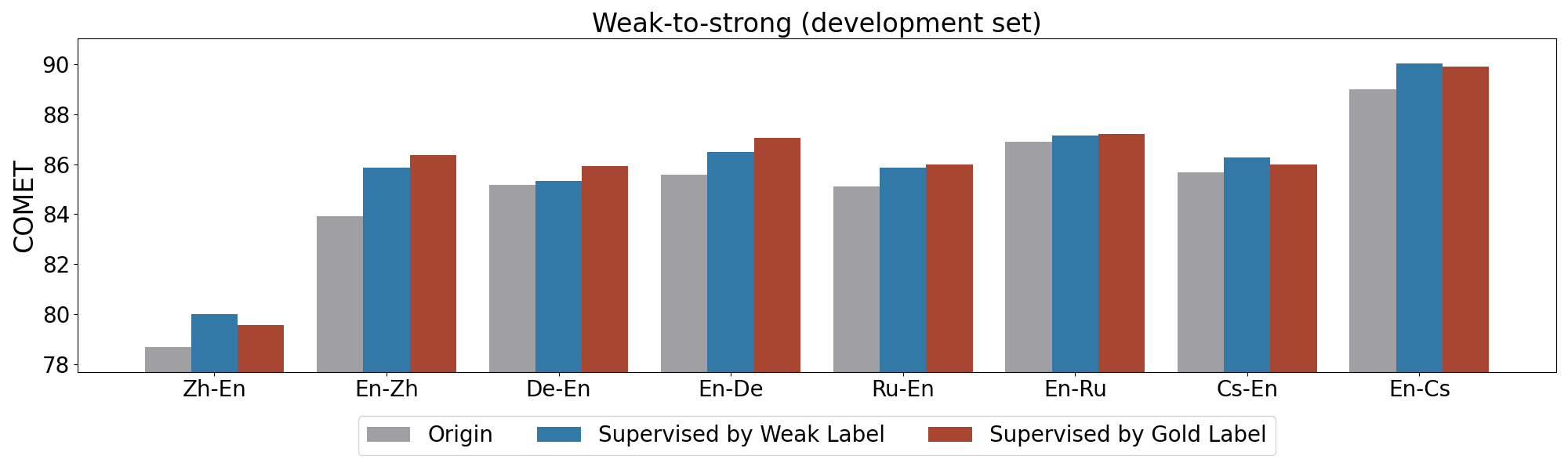}
    \caption{Weak-to-strong potential. We fine-tune Gemma-7B using different \textit{references} as the labels to refine the development set. \textcolor{gray}{Origin} denotes ALMA-7B-LoRA translation. \textcolor{deepblue}{Blue} represents using ALMA-7B-LoRA as the \textit{weak reference} to fine-tune MT-Ladder. \textcolor{deepred}{Red} represents using the gold label as the \textit{reference}.} 
    \vspace{-3mm}
    \label{fig:weak-comet}
\end{figure*}

\begin{figure*}[ht]
    \centering
    \includegraphics[scale=0.28]{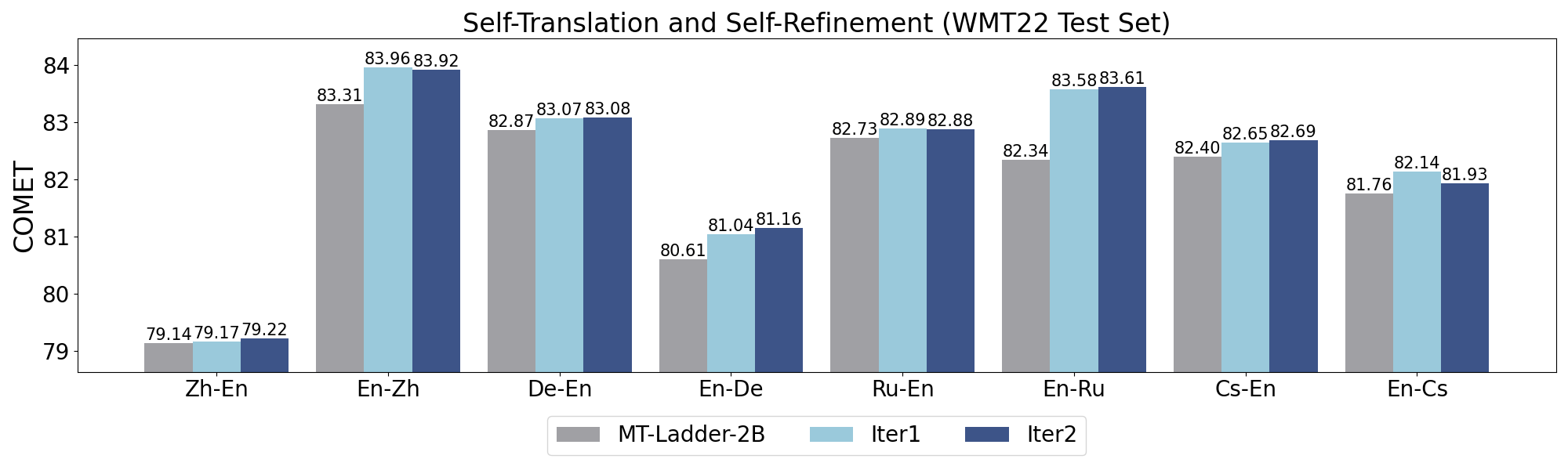}
    \caption{Self-translation and Self-refinement. \textcolor{gray}{MT-Ladder-2B} represents performing direct translation with prompt $\mathcal{P}_{D}$, demonstrating translation capabilities comparable to 7B and 13B LLM-based translators. \textcolor{iterblue1}{Iter1} denotes MT-Ladder-2B refining its original translation. \textcolor{iterblue2}{Iter2} denotes MT-Ladder-2B refining the translation from \textcolor{iterblue1}{Iter1}.} 
    \vspace{-3mm}
    \label{fig:ladder2b-comet}
\end{figure*}

\subsection{Ablation and Analysis }

\noindent {\textbf{Analysis of HFT.}} 
\label{sec:hierarchy_ft_effect}
As depicted in Figure~\ref{fig:hierarchy} \footnote{We examine the effectiveness of HFT with the Gemma-7B on the development set (see Appendix~\ref{app:dataset}), automatically saving 10 checkpoints to calculate metric scores.}, our HFT exhibits stable improvements and the best performance regarding BLEU and COMET in all ten checkpoints, while the traditional mixed training strategy fluctuates with inferior performance.  We also conduct another "Anti-HFT" experiment by reverting the order of the corpus employed during HFT, i.e., MT-Ladder is trained following a hard-to-easy schema. Results in Figure~\ref{fig:hierarchy} shows that "Anti-HFT" initially achieves its best performance and then gradually declines. 

We further scrutinize the model performance during HFT to verify its effectiveness. We report two metrics, the average improvement $\Delta$ and its standard deviation $\sigma$ of the above three strategies during the training process, while larger $\Delta$ and smaller $\sigma$ indicate better and more stable refinement improvements.  The results are in Figure~\ref{fig:anti_trend}. 

We notice that 
HFT results in a gradual increase of $\Delta$ and a decrease of $\sigma$. 
However, "Anti-HFT" shows the opposite trend, and the mixed training fluctuates in both $\Delta$ and $\sigma$. The increasing $\sigma$ in "Anti-HFT" suggests that learning on \textit{Easy} triplets might affect the stability of refinements.  These results align with our hypothesis that refining \textit{Hard} samples requires fewer adjustments, while \textit{Easy} samples, which exhibit substantial deviations from the reference, demand more corrections and can cause significant fluctuations if utilized for finetuning in the final stage. See samples in Table~\ref{tab:stage_case} and~\ref{tab:hierarchy_case} for intuitive understandings. Our findings suggest that the way triplet data is partitioned and ordered for HFT can impact model performance for instruction-following refinement, while more robust fine-tuning strategies are of high necessity in future work. 




We also investigate the sensitivity of the threshold $\mu$ and $\nu$ used for splitting hierarchies and conduct HFT with three different thresholds on En-Zh training set, as shown in Figure~\ref{fig:score_stages}. The results indicate that HFT consistently outperforms mixed training, with similar performance across different thresholds. 


\noindent {\textbf{Refinements Degrade as the Original LLM Becomes Stronger.}} We analyze the quality score changes between the original translations and the MT-Ladder-refined versions as shown in Figure~\ref{fig:trend}. We observe that MT-Ladder consistently improves a higher proportion of translations than it degrades, even for GPT-4. The trend in the proportion of improved translations aligns with the average score improvement trend. Specifically, as the model's translation capability increases, the proportion of improvements decreases, and the average improvement score also decreases. Our findings suggest that stronger translations have fewer and more complex errors that are harder to refine, consistent with our assumption in Section~\ref{sec:hierarchical optimization}.


\input{table/change_base}

\noindent {\textbf{Ablation Study of Different Sampling and Backbones.}}  
As shown in Table~\ref{tab:change_base}, MT-Ladder trained using different sampling and backbones consistently improves translation quality across instruction-tuning models of various sizes, demonstrating the effectiveness of our instruction-following refinement strategy. Notably, Gemma-2B (Vicuna-7B) with MT-Ladder even surpasses Gemma-7B (Vicuna-13B), highlighting the potential to enhance the capabilities of smaller models to next level.

\noindent {\textbf{Instruction-following Refinement Enables Weak-to-Strong Generalization.}} Typically, the capabilities after fine-tuning are upper-bounded by the supervised label, i.e., the \textit{reference} in our task. Here, we explore using ALMA-7B-LoRA sampled translation as the \textit{weak reference} and Vicuna-7B sampled translation as the \textit{intermediate translation} to create pseudo-refinement training triplets [\textit{source}, \textit{intermediate translation}, \textit{weak reference}]. Figure \ref{fig:weak-comet} and \ref{fig:weak-bleu} show that MT-Ladder trained under this weak supervision can refine translations from the weak label annotator ALMA-7B-LoRA, surpassing it in both BLEU and COMET scores. Remarkably, it even outperforms  gold label supervision in three translation directions. This demonstrates the potential of our instruction-following refinement method  to exceed the current limits of supervision.


\noindent {\textbf{MT-Ladder Can Act as a Good Translator and Execute Self-refinement.}} 
\label{sec:self-refine}
We evaluate the translation capability of MT-Ladder and explore its self-refinement potential. Figure~\ref{fig:ladder2b-comet} shows that MT-Ladder-2B can also execute the direct translation task and can improve its own initial translations across 8 translation directions, with increased COMET scores. However, the refinement effect becomes less pronounced with each iteration. More metrics are in Appendix~\ref{app:self-refine}.

\section{Related Work}
\paragraph{Automatic Post-Edition and Refinement} 
APE aims to cope with systematic errors of an MT system and adapt the output to the lexicon/style requested in a specific application domain. \citet{Correia2019} proposed a BERT-based method for APE using transfer learning. Other studies \citep{negri2018escape,vu-haffari-2018-automatic, chatterjee2019automatic,  Shterionov2019road, voita-etal-2019-context,gois2020learning, chollampatt2020pedra, Carmo2020ARO} investigated dataset construction, model architectures, and context integration to improve post-edited translations. 

With the development of LLMs, learning-based approaches have trained LLMs for refining translations to improve the overall translation segment quality \citep{xu2023pinpoint, alves2024tower, koneru2023contextual}. Recent works \citep{chen2023iterative, raunak2023leveraging, feng2024improving} have also explored using powerful LLMs, such as ChatGPT, to refine translations through prompting strategies like in-context learning and self-correction.

\paragraph{LLMs for Machine Translation} 
LLM-based machine translation falls into two main categories. The first focuses on strategies like prompt design, in-context example selection, and evaluation in various contexts such as low-resource, document-level, and multilingual translation \citep{vilar2022prompting, zhang2023prompting, peng-etal-2023-towards, wang2023document,liang2023encouraging, he2024exploring}. The second category focuses on training translation-specific LLMs. Prior studies \citep{zeng2023tim, jiao-etal-2023-parrot, kudugunta2024madlad, zan2024building, li2024eliciting, guo2024novel, he2024improving, wu2024adapting, xu2024contrastive} have explored aspects such as dataset construction, training paradigms, and exploring different contexts to achieve better translation performance.



\section{Conclusion}
In this paper, we introduce MT-Ladder, a model-agnostic and cost-effective tool for multilingual translation refinement that bridges the gap between off-the-shelf models and top-tier translation models. We sample translations from existing models to create pseudo-refinement training triplets without human annotations, which makes training cost-efficient. The proposed hierarchical fine-tuning strategy improves MT-Ladder's refining performance step by step, following an easy-to-hard schema. Our  exploration of training paradigms demonstrates good performance in effectiveness and robustness, as well as promising results in weak-to-strong generalization and self-refinement, providing valuable insights to the MT area.

\section*{Limitations}
Although MT-Ladder has shown promising results in bridging the gap between the translation performance of different models, it has some limitations. We have validated MT-Ladder's support for sentence-level translations, but document-level support still needs exploration. Expanding MT-Ladder's usage to support more languages, especially low-resource languages, is also crucial for future work. Additionally, deploying this approach to larger models (e.g., 70B) or smaller models (e.g., less than 1B) is worth exploring in future research. Leveraging the principles of MT-Ladder to explore instruction-following refinement in more generation tasks is also an interesting direction for future work.

\section*{Acknowledgments}
This work is supported by the National Natural Science Foundation of China (Grant No. 62106222), the Natural Science Foundation of Zhejiang Province, China (Grant No. LZ23F020008), and the Zhejiang University-Angelalign Inc. R\&D Center for Intelligent Healthcare.


\bibliography{custom}

\appendix

\section{Dataset Statistics}
\label{app:dataset}
Table~\ref{tab:data} presents statistic details of the data we used. For the development set, we randomly sampled 100 examples from the development parallel data and used ALMA-7B-LoRA to generate intermediate translations, totaling 800 development triplets.

\section{Baseline Models}
\label{app:Basemodel}
\paragraph{Translation Models} 
\begin{itemize}
    \item BigTranslate \citep{yang-etal-2023-BigTranslate} extends LLaMA to over 100 translation directions.
    \item BayLing \citep{zhang2023bayling} is an instruction-following large language model equipped with advanced language alignment.
    \item NLLB \citep{costa2022no} is a translation model with encoder-decoder architecture.
    \item ALMA \citep{xu2023paradigm} is a many-to-many LLM-based translation model. It represents the top level of open-source translators.
\end{itemize}

\paragraph{Non-translation Models}
\begin{itemize}
    \item Alpaca \citep{alpaca} is a LLaMA Model fine-tuned on 52K instruction-following data.
    \item Vicuna-v1.5 \citep{vicuna2023} is fine-tuned from LLaMA2 with supervised instruction fine-tuning. The training data is around 125K conversations collected from ShareGPT~\footnote{\href{https://sharegpt.com}{https://sharegpt.com}}.  
    \item text-davinci-003 is a GPT-3.5 model with 175B parameters \citep{ouyang2022training}.
    \item GPT-4 \citep{achiam2023gpt} is the latest and the most powerful version of GPT-series. We use OpenAI API gpt-4-1106-preview. 
\end{itemize}

\paragraph{SoTA APE Models}
\begin{itemize}
    \item Contrast Translation (CT) and Rephrase (Re) \citep{chen2023iterative} are two prompt-based translation refinement methods. \textit{CT} means inserting the word "bad" in the prompts to ask the instruction-following LLM do the contrastive translation. \textit{Re} refers to asking the LLM to rephrase the original translation.

    \item LLMRefine \citep{xu2023pinpoint} is fine-tuned on PaLM2 (Bison) to refine LLM’s output with fine-grained actionable feedback iteratively.
    
    \item TowerInstruct \citep{alves2024tower} is an effective translation post editor. It is fine-tuned on high-quality parallel translation data totaling 637k examples. The APE-related tasks include MQM evaluation data (WMT20 to WMT22) annotated with multidimensional quality metrics~\citep{freitag2021experts}, accounting for 20.9\%. Translation data with post-edits from QT21~\citep{specia-etal-2017-translation} and ApeQuest~\footnote{\href{https://apequest.wordpress.com/}{https://apequest.wordpress.com/}} are used for automatic post-editing, making up 3.1\% and 3.3\% of the data, respectively. TowerInstruct outperforms open models and GPT-3.5-turbo on APE.
    
    \item GPT-4o mini scores 82\% on MMLU and currently outperforms GPT-4-turbo-0125 on chat preferences in LMSYS leaderboard. GPT-4o mini surpasses GPT-3.5 Turbo and other small models on academic benchmarks and supports the same range of languages as GPT-4o~\footnote{\href{https://openai.com/index/gpt-4o-mini-advancing-cost-efficient-intelligence/}{https://openai.com/index/gpt-4o-mini-advancing-cost-efficient-intelligence/}}.
\end{itemize}

\section{Training Details}
\label{app:details}
We fine-tune our model using LoRA with a rank of 16 and a learning rate of 1e-4. All models are fine-tuned for 1 epoch with a batch size of 16, imposing a maximum text length of 512. We adopt deepspeed \citep{deepspeed} to accelerate our training. 

\section{Base Model Effect}
\begin{figure*}[ht]
    \centering
    \includegraphics[scale=0.30]{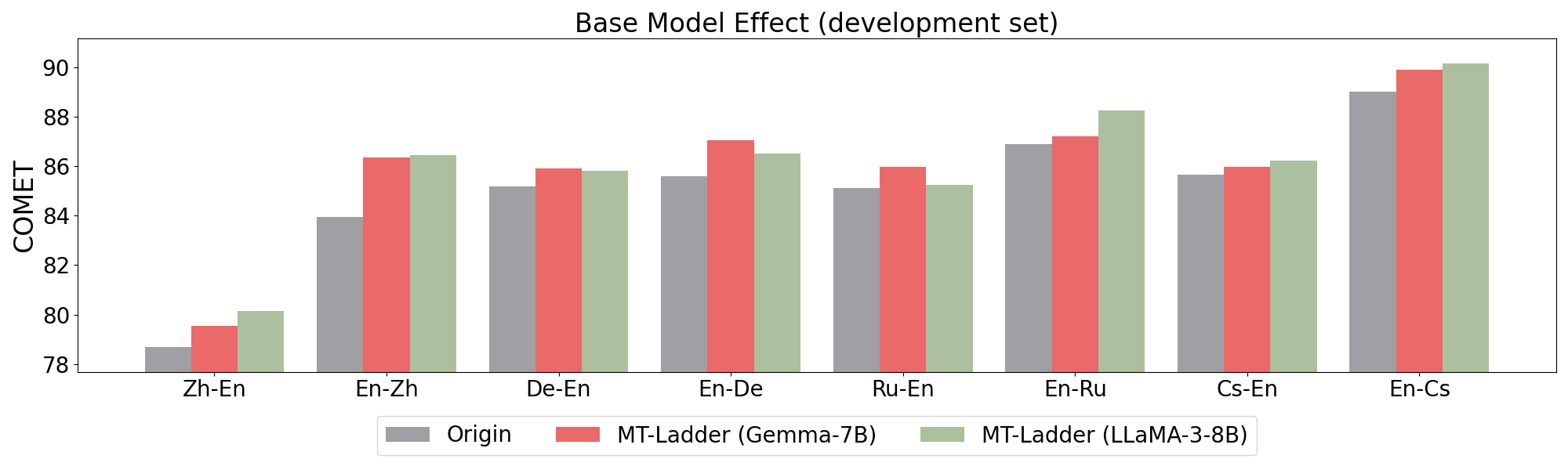}
    \caption{COMET scores of using different models as the base model of MT-Ladder. \textcolor{gray}{Origin} denotes ALMA-7B-LoRA translation.} 
    \vspace{-3mm}
    \label{fig:base-model-effect}
\end{figure*}

We also finetuned LLaMA-3-8B\footnote{\href{https://github.com/meta-llama/llama3}{https://github.com/meta-llama/llama3}} using the proposed HFT method on the same training set as MT-Ladder-7B and evaluated it on the development set to refine ALMA-7B-LoRA. The results in Figure~\ref{fig:base-model-effect} 
shows that In the XX-En direction, LLaMA-3-8B achieved higher scores in Zh-En and Cs-En but lagged behind Gemma-7B in the other two translation directions, resulting in comparable average scores between the two models. In the En-XX direction, LLaMA-3-8B outperformed Gemma-7B in three out of four translation directions, with a higher average score overall compared to the current MT-Ladder-7B. This observation suggests that LLaMA-3-8B's enhanced multilingual capabilities, inherent to its pre-training phase, benefited from exposure to a broader multilingual dataset. We consider the selection of the base model as a crucial direction for future improvements to MT-Ladder.

\section{Self-translation and Self-refinement}
\label{app:self-refine}
\begin{figure*}[ht]
    \centering
    \includegraphics[scale=0.30]{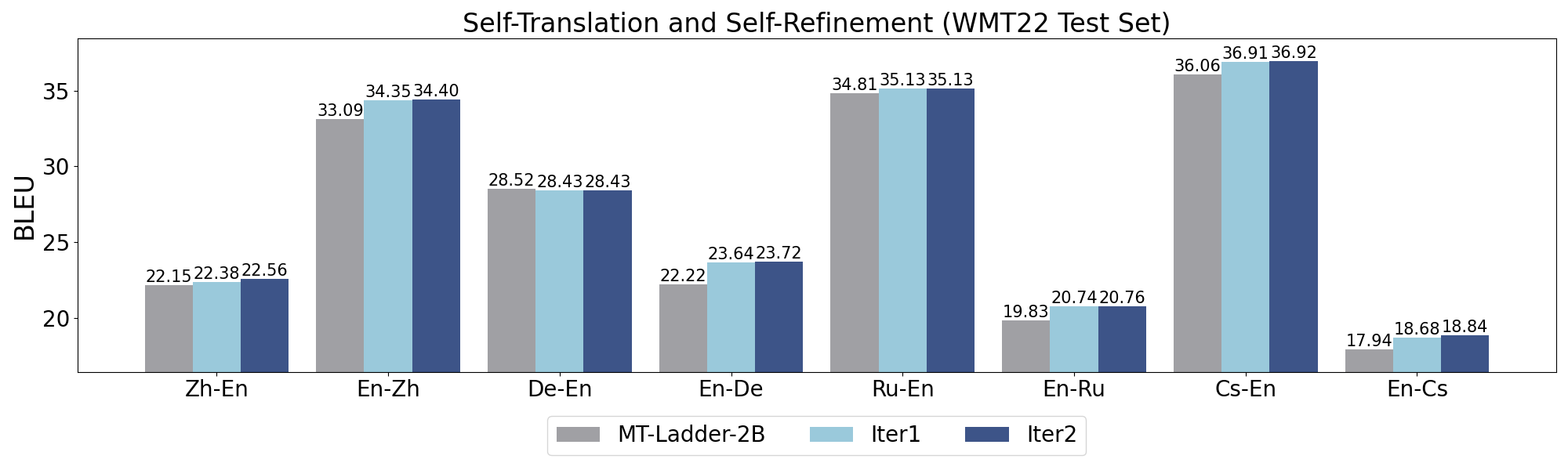}
    \caption{BLEU scores for Self-translation and Self-refinement. \textcolor{iterblue1}{Iter1} denotes MT-Ladder-2B refines its original translation. \textcolor{iterblue2}{Iter2} denotes MT-Ladder-2B refines the MT-Ladder-2B edited translation in \textcolor{iterblue1}{Iter1}.} 
    \vspace{-3mm}
    \label{fig:ladder2b-bleu}
\end{figure*}

\begin{figure*}[ht]
    \centering
    \includegraphics[scale=0.30]{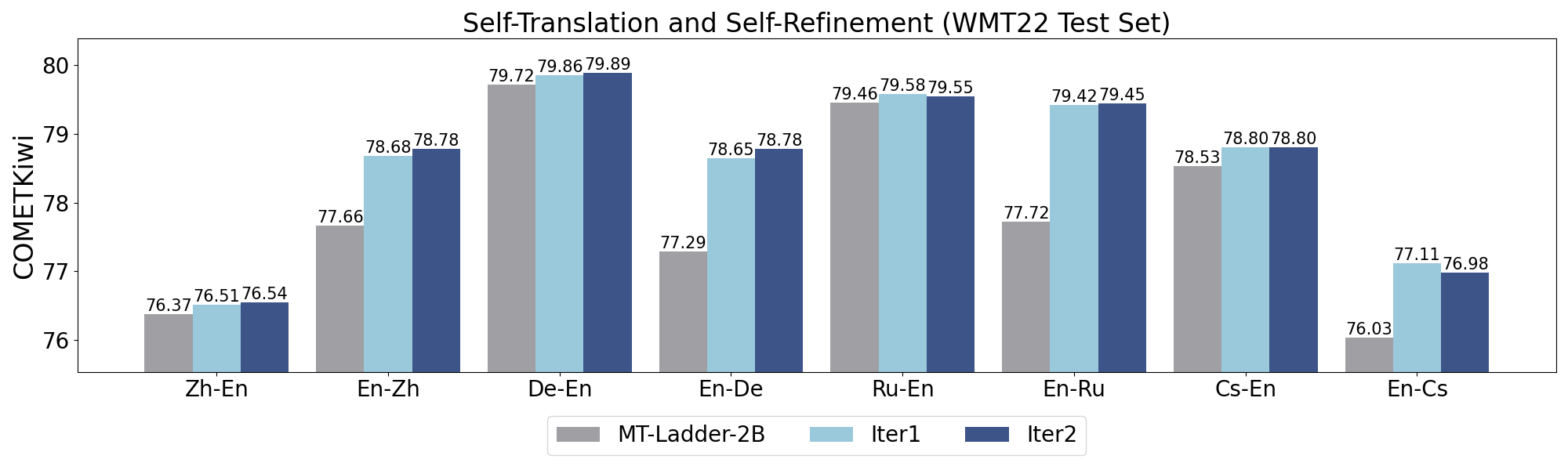}
    \caption{COMETKiwi scores for Self-translation and Self-refinement. \textcolor{iterblue1}{Iter1} denotes MT-Ladder-2B refines its original translation. \textcolor{iterblue2}{Iter2} denotes MT-Ladder-2B refines the MT-Ladder-2B edited translation in \textcolor{iterblue1}{Iter1}.} 
    \vspace{-3mm}
    \label{fig:ladder2b-cometkiwi}
\end{figure*}

\begin{figure*}[ht]
    \centering
    \includegraphics[scale=0.30]{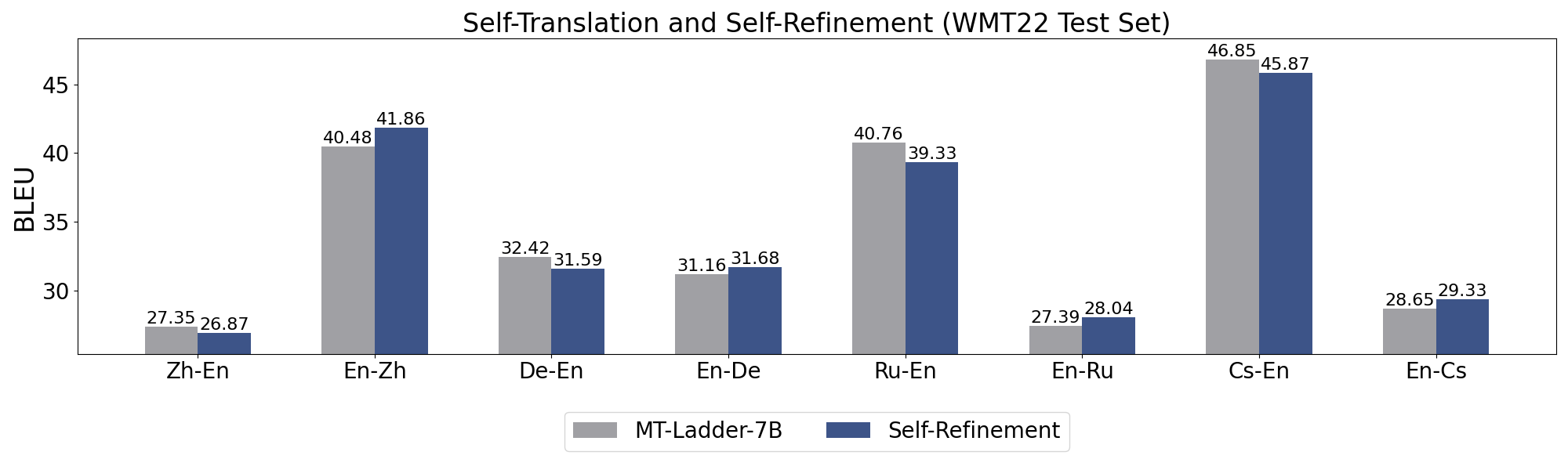}
    \caption{BLEU scores for Self-translation and Self-refinement with MT-Ladder-7B. \textcolor{iterblue2}{Self-Refinement} denotes MT-Ladder-7B refines its original translation.} 
    \vspace{-3mm}
    \label{fig:ladder7b-bleu}
\end{figure*}

\begin{figure*}[ht]
    \centering
    \includegraphics[scale=0.30]{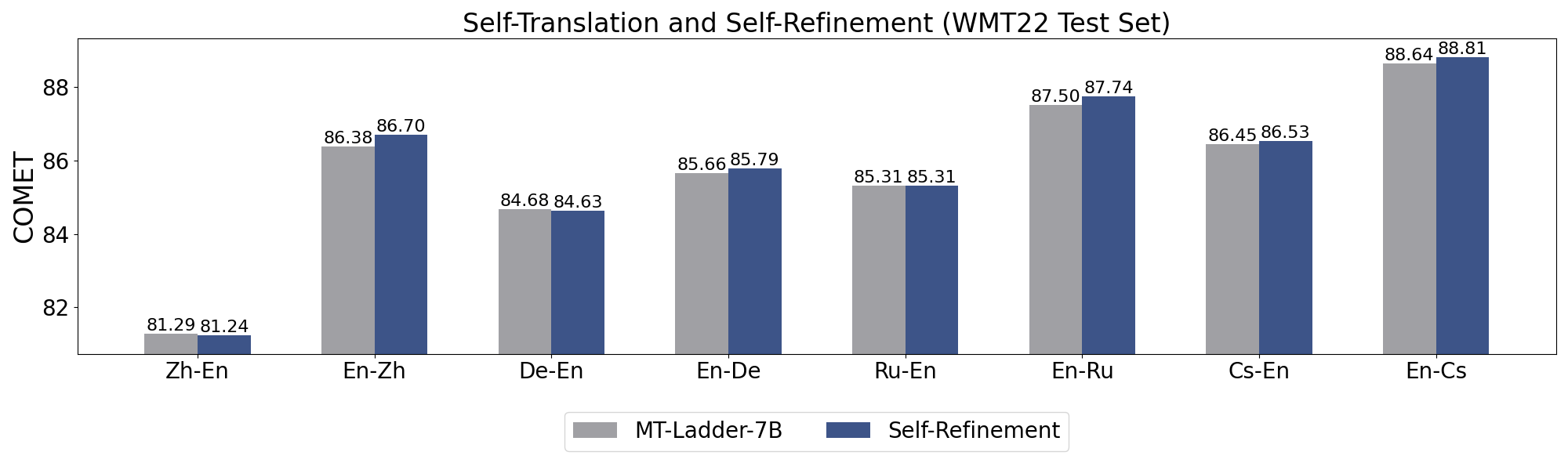}
    \caption{COMET scores for Self-translation and Self-refinement with MT-Ladder-7B. \textcolor{iterblue2}{Self-Refinement} denotes MT-Ladder-7B refines its original translation.} 
    \vspace{-3mm}
    \label{fig:ladder7b-comet}
\end{figure*}

\begin{figure*}[ht]
    \centering
    \includegraphics[scale=0.30]{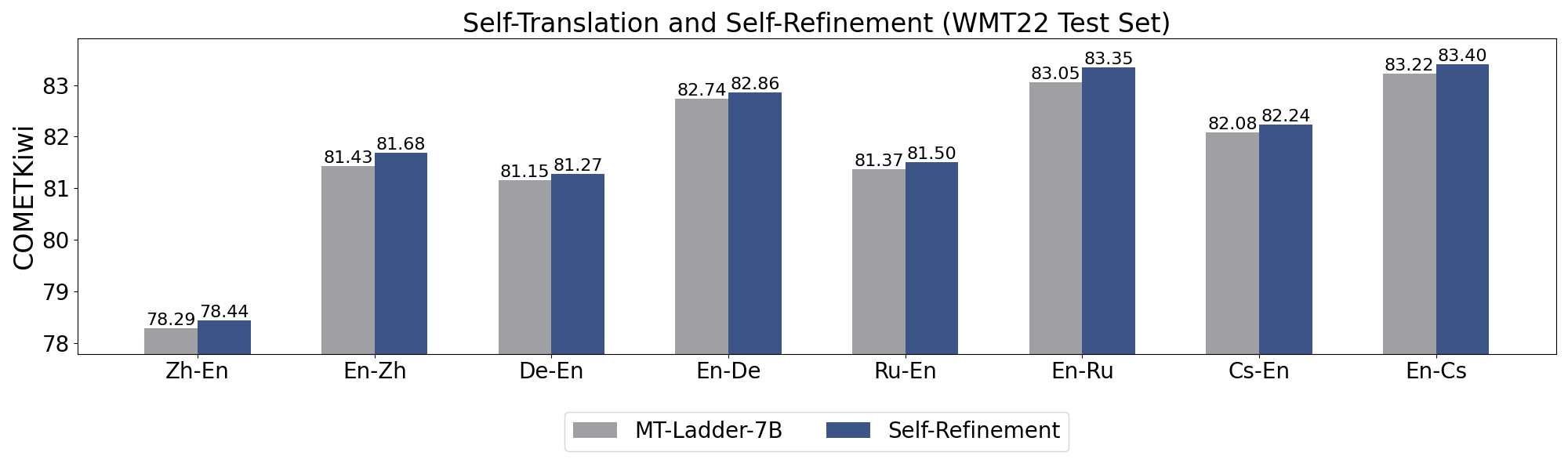}
    \caption{COMETKiwi scores for Self-translation and Self-refinement with MT-Ladder-7B. \textcolor{iterblue2}{Self-Refinement} denotes MT-Ladder-7B post-edits its original translation.} 
    \vspace{-3mm}
    \label{fig:ladder7b-cometkiwi}
\end{figure*}

For Section~\ref{sec:self-refine}, we supplement the BLEU and COMETKiwi of MT-Ladder-2B (see Figure~\ref{fig:ladder2b-bleu} and~\ref{fig:ladder2b-cometkiwi}) and all metrics of MT-Ladder-7B (see Figure~\ref{fig:ladder7b-bleu}, \ref{fig:ladder7b-comet} and \ref{fig:ladder7b-cometkiwi}).

\begin{figure*}[ht]
    \centering
    \includegraphics[scale=0.3]{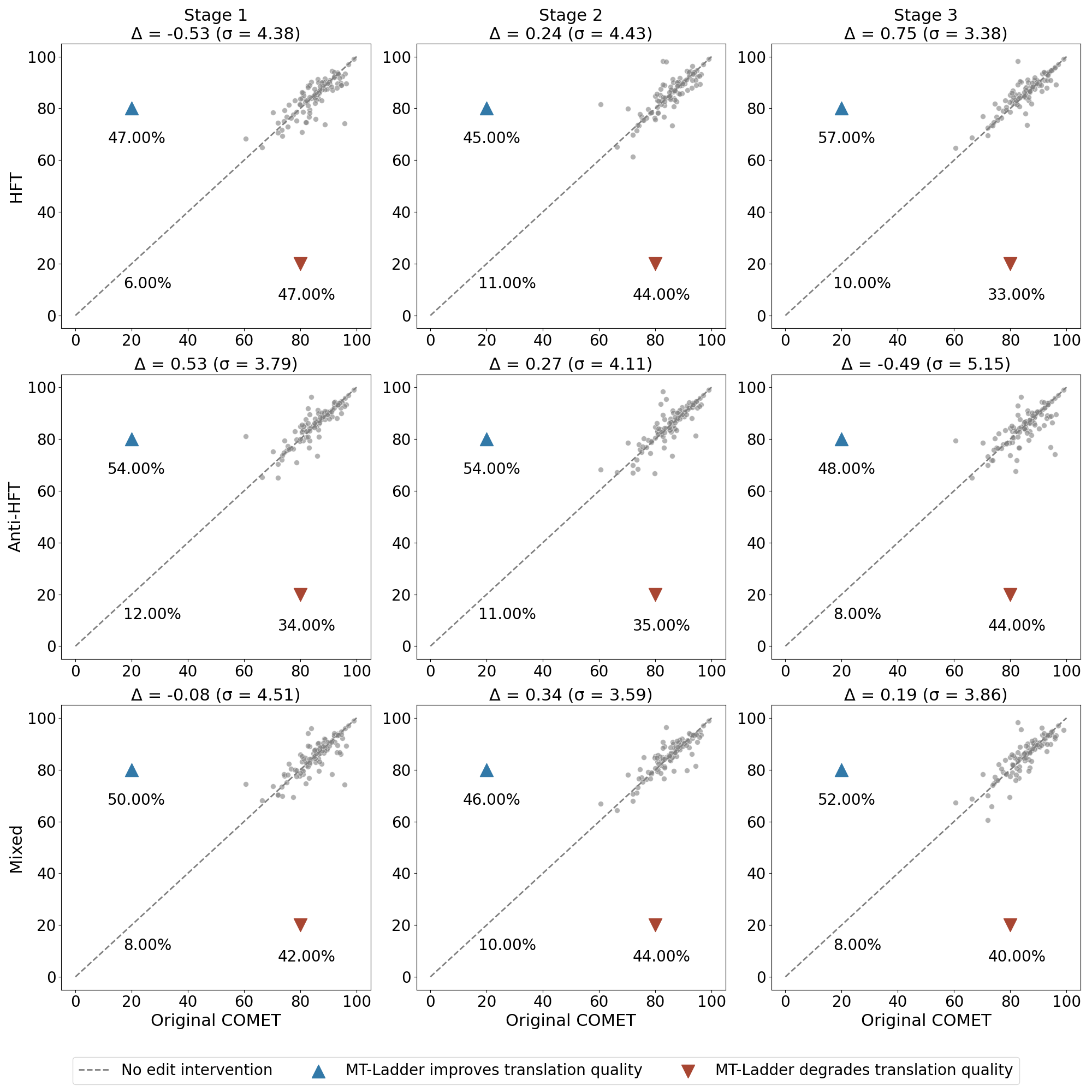}
    \caption{Comparison of original quality (x-axis) with refined quality (y-axis) in different fine-tuning stages. Each dot is a WMT22 De-En translation in our development set. We select the checkpoint at 2, 6, and 10 from Figure~\ref{fig:hierarchy} (which we refer to as Stage 1, Stage 2 and Stage 3 here). $\Delta$ denotes the average improvement. $\sigma$ refers to the standard deviation of $\Delta$. The percentages represent the proportion of each part, attached next to the markers.} 
    \vspace{-5mm}
    \label{fig:anti_trend}
\end{figure*}

\input{appendix/data_parallel}

\begin{figure*}[ht]
    \centering
    \includegraphics[scale=0.30]{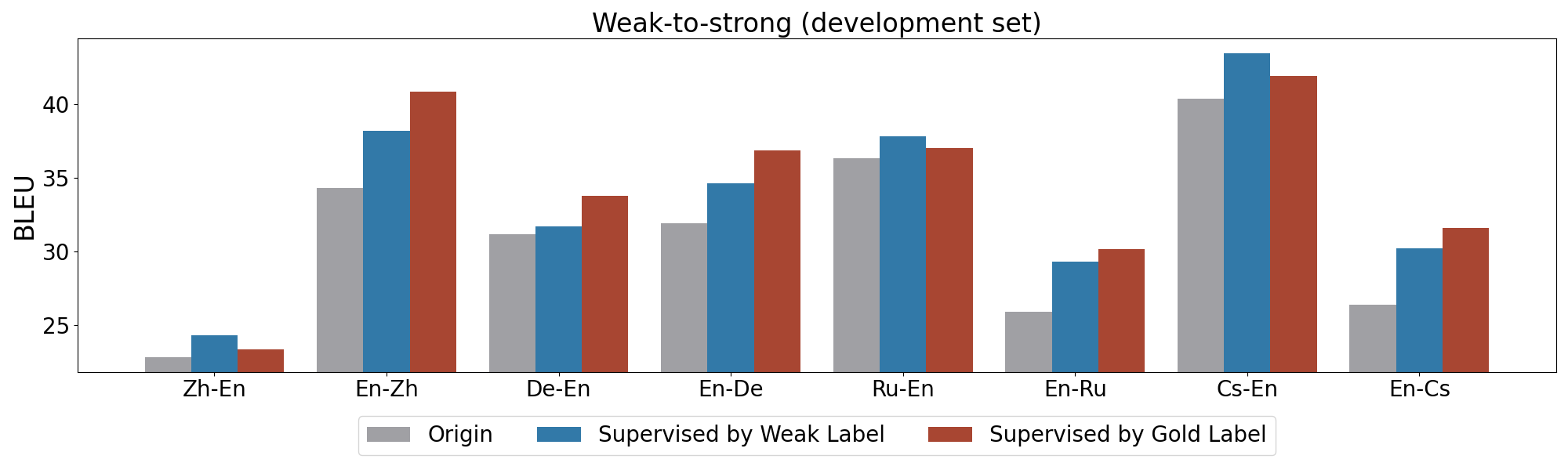}
    \caption{Weak-to-strong BLEU scores. We fine-tune Gemma-7B using different \textit{references} as the label to refine the development set.  \textcolor{gray}{Origin} denotes ALMA-7B-LoRA translation. \textcolor{deepblue}{Blue} represents using ALMA-7B-LoRA as \textit{references}. \textcolor{deepred}{Red} represents using the gold as \textit{references}.} 
    \vspace{-3mm}
    \label{fig:weak-bleu}
\end{figure*}

\input{appendix/hft_stage_case}
\input{appendix/hierarchical_case}

\label{sec:appendix}

\end{document}

%% file: table/xxen_all.tex



\begin{table*}[ht]
\centering
\resizebox{\textwidth}{!}{%
\begin{tabular}{lcclcclcclcclcc} 
\toprule
\multirow{2}{*}{\textbf{Models}} & \multicolumn{2}{c}{\textbf{\textbf{Zh-En}}} & \multicolumn{1}{c}{} & \multicolumn{2}{c}{\textbf{De-En}} & \multicolumn{1}{c}{} & \multicolumn{2}{c}{\textbf{\textbf{\textbf{\textbf{Ru-En}}}}} &  & \multicolumn{2}{c}{\textbf{\textbf{\textbf{\textbf{Cs-En}}}}} & \multicolumn{1}{c}{} & \multicolumn{2}{c}{\textbf{Avg.}} \\ 
\cmidrule{2-3}\cmidrule{5-6}\cmidrule{8-9}\cmidrule{11-12}\cmidrule{14-15}
 & \textbf{BLEU} & \textbf{COMET} &  & \textbf{BLEU} & \textbf{COMET} &  & \textbf{BLEU} & \textbf{COMET} &  & \textbf{BLEU} & \textbf{COMET} &  & \textbf{BLEU} & \textbf{COMET} \\ 
\midrule
\multicolumn{15}{c}{\textit{Open}} \\
Alpaca-7B & 11.80 & 73.36 &  & 24.52 & 81.37 &  & 30.49 & 80.68 &  & 27.31 & 77.99 &  & 23.53 & 78.35 \\
BigTranslate-13B & 14.32 & 74.63 &  & 23.17 & 81.04 &  & 28.05 & 78.38 &  & 34.49 & 81.99 &  & 25.01 & 79.01 \\
BayLing-13B & 20.12 & 77.72 &  & 27.36 & 83.03 &  & 33.95 & 82.07 &  & 33.87 & 81.64 &  & 28.83 & 81.12 \\
Vicuna-7B-v1.5 & 19.99  & 78.97 &  & 28.96 & 83.38 &  & 35.06 & 82.54 &  & 34.56 & 81.71 &  & 29.64 & 81.65 \\
NLLB-3.3B & 21.07 & 76.93 &  & 29.55 & 83.43 &  & 40.08 & 83.95 &  & 49.06 & 85.92 &  & 34.94 & 82.56 \\
ALMA-7B-LoRA & 24.00 & 80.18 &  & 29.98 & 84.16 &  & 38.43 & 84.80 &  & 43.96 & 86.00 &  & 34.09 & 83.79 \\
ALMA-13B-LoRA & 25.48 & 80.21 &  & 31.26 & 84.56 &  & 40.26 & 85.27 &  & 45.36 & 86.47 &  & 35.59 & 84.13 \\ 
\hdashline
\multicolumn{15}{c}{\textit{Closed}} \\
text-davinci-003 & 25.00 & 81.62 &  & 30.88 & 84.79 &  & 38.47 & 84.80 &  & 44.52 & 86.16 &  & 34.72 & 84.34 \\
GPT-4 & 23.80 & 82.46 &  & 32.46 & 85.35 &  & 40.98 & 85.87 &  & 46.77 & 87.26 &  & 36.00 & 85.24 \\ 
\midrule
\multicolumn{15}{c}{\textit{MT-Ladder-2B Refinement}}                                                                                                                                                                                                                                                         \\
\vcell{Alpaca-7B} & \vcell{\begin{tabular}[b]{@{}c@{}}22.73\\\cellcolor{lightblue}(+10.93)\end{tabular}} & \vcell{\begin{tabular}[b]{@{}c@{}}78.98\\\cellcolor{lightblue}(+5.62)\end{tabular}} & \vcell{} & \vcell{\begin{tabular}[b]{@{}c@{}}28.53\\\cellcolor{lightblue}(+4.01)\end{tabular}} & \vcell{\begin{tabular}[b]{@{}c@{}}83.34\\\cellcolor{lightblue}(+1.97)\end{tabular}} & \vcell{} & \vcell{\begin{tabular}[b]{@{}c@{}}36.05\\\cellcolor{lightblue}(+5.56)\end{tabular}} & \vcell{\begin{tabular}[b]{@{}c@{}}83.34\\\cellcolor{lightblue}(+2.66)\end{tabular}} & \vcell{} & \vcell{\begin{tabular}[b]{@{}c@{}}37.08\\\cellcolor{lightblue}(+9.77)\end{tabular}} & \vcell{\begin{tabular}[b]{@{}c@{}}83.08\\\cellcolor{lightblue}(+5.09)\end{tabular}} & \vcell{} & \vcell{\begin{tabular}[b]{@{}c@{}}31.10\\\cellcolor{lightblue}(+7.57)\end{tabular}} & \vcell{\begin{tabular}[b]{@{}c@{}}82.19\\\cellcolor{lightblue}(+3.84)\end{tabular}} \\[-\rowheight]
\printcelltop & \printcellmiddle & \printcellmiddle & \printcellmiddle & \printcellmiddle & \printcellmiddle & \printcellmiddle & \printcellmiddle & \printcellmiddle & \printcellmiddle & \printcellmiddle & \printcellmiddle & \printcellmiddle & \printcellmiddle & \printcellmiddle \\
\vcell{BigTranslate-13B} & \vcell{\begin{tabular}[b]{@{}c@{}}22.58\\\cellcolor{lightblue}(+8.26)\end{tabular}} & \vcell{\begin{tabular}[b]{@{}c@{}}79.28\\\cellcolor{lightblue}(+4.65)\end{tabular}} & \vcell{} & \vcell{\begin{tabular}[b]{@{}c@{}}28.48\\\cellcolor{lightblue}(+5.31)\end{tabular}} & \vcell{\begin{tabular}[b]{@{}c@{}}83.45\\\cellcolor{lightblue}(+2.41)\end{tabular}} & \vcell{} & \vcell{\begin{tabular}[b]{@{}c@{}}36.31\\\cellcolor{lightblue}(+8.26)\end{tabular}} & \vcell{\begin{tabular}[b]{@{}c@{}}83.22\\\cellcolor{lightblue}(+4.84)\end{tabular}} & \vcell{} & \vcell{\begin{tabular}[b]{@{}c@{}}40.32\\\cellcolor{lightblue}(+5.83)\end{tabular}} & \vcell{\begin{tabular}[b]{@{}c@{}}84.15\\\cellcolor{lightblue}(+2.16)\end{tabular}} & \vcell{} & \vcell{\begin{tabular}[b]{@{}c@{}}31.92\\\cellcolor{lightblue}(+6.91)\end{tabular}} & \vcell{\begin{tabular}[b]{@{}c@{}}82.53\\\cellcolor{lightblue}(+3.52)\end{tabular}} \\[-\rowheight]
\printcelltop & \printcellmiddle & \printcellmiddle & \printcellmiddle & \printcellmiddle & \printcellmiddle & \printcellmiddle & \printcellmiddle & \printcellmiddle & \printcellmiddle & \printcellmiddle & \printcellmiddle & \printcellmiddle & \printcellmiddle & \printcellmiddle \\
\vcell{BayLing-13B} & \vcell{\begin{tabular}[b]{@{}c@{}}23.84\\\cellcolor{lightblue}(+3.72)\end{tabular}} & \vcell{\begin{tabular}[b]{@{}c@{}}79.55\\\cellcolor{lightblue}(+1.83)\end{tabular}} & \vcell{} & \vcell{\begin{tabular}[b]{@{}c@{}}29.05\\\cellcolor{lightblue}(+1.69)\end{tabular}} & \vcell{\begin{tabular}[b]{@{}c@{}}83.64\\\cellcolor{lightblue}(+0.61)\end{tabular}} & \vcell{} & \vcell{\begin{tabular}[b]{@{}c@{}}36.92\\\cellcolor{lightblue}(+2.97)\end{tabular}} & \vcell{\begin{tabular}[b]{@{}c@{}}83.69\\\cellcolor{lightblue}(+1.62)\end{tabular}} & \vcell{} & \vcell{\begin{tabular}[b]{@{}c@{}}38.85\\\cellcolor{lightblue}(+4.98)\end{tabular}} & \vcell{\begin{tabular}[b]{@{}c@{}}83.59\\\cellcolor{lightblue}(+1.95)\end{tabular}} & \vcell{} & \vcell{\begin{tabular}[b]{@{}c@{}}32.17\\\cellcolor{lightblue}(+3.34)\end{tabular}} & \vcell{\begin{tabular}[b]{@{}c@{}}82.61\\\cellcolor{lightblue}(+1.49)\end{tabular}} \\[-\rowheight]
\printcelltop & \printcellmiddle & \printcellmiddle & \printcellmiddle & \printcellmiddle & \printcellmiddle & \printcellmiddle & \printcellmiddle & \printcellmiddle & \printcellmiddle & \printcellmiddle & \printcellmiddle & \printcellmiddle & \printcellmiddle & \printcellmiddle \\
\vcell{Vicuna-7B-v1.5} & \vcell{\begin{tabular}[b]{@{}c@{}}24.11\\\cellcolor{lightblue}(+4.12)\end{tabular}} & \vcell{\begin{tabular}[b]{@{}c@{}}80.05\\\cellcolor{lightblue}(+1.08)\end{tabular}} & \vcell{} & \vcell{\begin{tabular}[b]{@{}c@{}}29.85\\\cellcolor{lightblue}(+0.89)\end{tabular}} & \vcell{\begin{tabular}[b]{@{}c@{}}83.76\\\cellcolor{lightblue}(+0.38)\end{tabular}} & \vcell{} & \vcell{\begin{tabular}[b]{@{}c@{}}37.72\\\cellcolor{lightblue}(+2.66)\end{tabular}} & \vcell{\begin{tabular}[b]{@{}c@{}}83.85\\\cellcolor{lightblue}(+1.31)\end{tabular}} & \vcell{} & \vcell{\begin{tabular}[b]{@{}c@{}}38.81\\\cellcolor{lightblue}(+4.25)\end{tabular}} & \vcell{\begin{tabular}[b]{@{}c@{}}83.60\\\cellcolor{lightblue}(+1.89)\end{tabular}} & \vcell{} & \vcell{\begin{tabular}[b]{@{}c@{}}32.62\\\cellcolor{lightblue}(+2.98)\end{tabular}} & \vcell{\begin{tabular}[b]{@{}c@{}}82.82\\\cellcolor{lightblue}(+1.17)\end{tabular}} \\[-\rowheight]
\printcelltop & \printcellmiddle & \printcellmiddle & \printcellmiddle & \printcellmiddle & \printcellmiddle & \printcellmiddle & \printcellmiddle & \printcellmiddle & \printcellmiddle & \printcellmiddle & \printcellmiddle & \printcellmiddle & \printcellmiddle & \printcellmiddle \\
\vcell{NLLB-3.3B} & \vcell{\begin{tabular}[b]{@{}c@{}}23.97\\\cellcolor{lightblue}(+2.90)\end{tabular}} & \vcell{\begin{tabular}[b]{@{}c@{}}79.34\\\cellcolor{lightblue}(+2.41)\end{tabular}} & \vcell{} & \vcell{\begin{tabular}[b]{@{}c@{}}29.83\\\cellcolor{lightblue}(+0.28)\end{tabular}} & \vcell{\begin{tabular}[b]{@{}c@{}}83.89\\\cellcolor{lightblue}(+0.46)\end{tabular}} & \vcell{} & \vcell{\begin{tabular}[b]{@{}c@{}}39.02\\\cellcolor{lightred}(-1.06)\end{tabular}} & \vcell{\begin{tabular}[b]{@{}c@{}}84.27\\\cellcolor{lightblue}(+0.32)\end{tabular}} & \vcell{} & \vcell{\begin{tabular}[b]{@{}c@{}}45.10\\\cellcolor{lightred}(-3.96)\end{tabular}} & \vcell{\begin{tabular}[b]{@{}c@{}}85.30\\\cellcolor{lightred}(-0.62)\end{tabular}} & \vcell{} & \vcell{\begin{tabular}[b]{@{}c@{}}34.48\\\cellcolor{lightred}(-0.46)\end{tabular}} & \vcell{\begin{tabular}[b]{@{}c@{}}83.20\\\cellcolor{lightblue}(+0.64)\end{tabular}} \\[-\rowheight]
\printcelltop & \printcellmiddle & \printcellmiddle & \printcellmiddle & \printcellmiddle & \printcellmiddle & \printcellmiddle & \printcellmiddle & \printcellmiddle & \printcellmiddle & \printcellmiddle & \printcellmiddle & \printcellmiddle & \printcellmiddle & \printcellmiddle \\ 
\midrule

\multicolumn{15}{c}{\textit{MT-Ladder-7B Refinement}} \\
\vcell{BigTranslate-13B} & \vcell{\begin{tabular}[b]{@{}c@{}}26.49\\\cellcolor{lightblue}(+12.17)\end{tabular}} & \vcell{\begin{tabular}[b]{@{}c@{}}81.08\\\cellcolor{lightblue}(+6.45)\end{tabular}} & \vcell{} & \vcell{\begin{tabular}[b]{@{}c@{}}31.13\\\cellcolor{lightblue}(+7.96)\end{tabular}} & \vcell{\begin{tabular}[b]{@{}c@{}}84.58\\\cellcolor{lightblue}(+3.54)\end{tabular}} & \vcell{} & \vcell{\begin{tabular}[b]{@{}c@{}}39.22\\\cellcolor{lightblue}(+11.17)\end{tabular}} & \vcell{\begin{tabular}[b]{@{}c@{}}85.25\\\cellcolor{lightblue}(+6.87)\end{tabular}} & \vcell{} & \vcell{\begin{tabular}[b]{@{}c@{}}45.87\\\cellcolor{lightblue}(+11.38)\end{tabular}} & \vcell{\begin{tabular}[b]{@{}c@{}}86.43\\\cellcolor{lightblue}(+4.44)\end{tabular}} & \vcell{} & \vcell{\begin{tabular}[b]{@{}c@{}}35.68\\\cellcolor{lightblue}(+10.67)\end{tabular}} & \vcell{\begin{tabular}[b]{@{}c@{}}84.34\\\cellcolor{lightblue}(+4.83)\end{tabular}} \\[-\rowheight]
\printcelltop & \printcellmiddle & \printcellmiddle & \printcellmiddle & \printcellmiddle & \printcellmiddle & \printcellmiddle & \printcellmiddle & \printcellmiddle & \printcellmiddle & \printcellmiddle & \printcellmiddle & \printcellmiddle & \printcellmiddle & \printcellmiddle \\
\vcell{NLLB-3.3B} & \vcell{\begin{tabular}[b]{@{}c@{}}26.91\\\cellcolor{lightblue}(+5.84)\end{tabular}} & \vcell{\begin{tabular}[b]{@{}c@{}}81.25\\\cellcolor{lightblue}(+4.32)\end{tabular}} & \vcell{} & \vcell{\begin{tabular}[b]{@{}c@{}}32.37\\\cellcolor{lightblue}(+2.82)\end{tabular}} & \vcell{\begin{tabular}[b]{@{}c@{}}84.88\\\cellcolor{lightblue}(+1.45)\end{tabular}} & \vcell{} & \vcell{\begin{tabular}[b]{@{}c@{}}41.97\\\cellcolor{lightblue}(+1.89)\end{tabular}} & \vcell{\begin{tabular}[b]{@{}c@{}}85.65\\\cellcolor{lightblue}(+1.70)\end{tabular}} & \vcell{} & \vcell{\begin{tabular}[b]{@{}c@{}}50.11\\\cellcolor{lightblue}(+1.05)\end{tabular}} & \vcell{\begin{tabular}[b]{@{}c@{}}87.09\\\cellcolor{lightblue}(+1.17)\end{tabular}} & \vcell{} & \vcell{\begin{tabular}[b]{@{}c@{}}37.84\\\cellcolor{lightblue}(+2.90)\end{tabular}} & \vcell{\begin{tabular}[b]{@{}c@{}}84.72\\\cellcolor{lightblue}(+2.16)\end{tabular}} \\[-\rowheight]
\printcelltop & \printcellmiddle & \printcellmiddle & \printcellmiddle & \printcellmiddle & \printcellmiddle & \printcellmiddle & \printcellmiddle & \printcellmiddle & \printcellmiddle & \printcellmiddle & \printcellmiddle & \printcellmiddle & \printcellmiddle & \printcellmiddle \\

\vcell{ALMA-7B-LoRA} & \vcell{\begin{tabular}[b]{@{}c@{}}26.91\\\cellcolor{lightblue}(+2.91)\end{tabular}} & \vcell{\begin{tabular}[b]{@{}c@{}}81.39\\\cellcolor{lightblue}(+1.21)\end{tabular}} & \vcell{} & \vcell{\begin{tabular}[b]{@{}c@{}}31.61\\\cellcolor{lightblue}(+1.63)\end{tabular}} & \vcell{\begin{tabular}[b]{@{}c@{}}84.65\\\cellcolor{lightblue}(+0.49)\end{tabular}} & \vcell{} & \vcell{\begin{tabular}[b]{@{}c@{}}39.42\\\cellcolor{lightblue}(+0.99)\end{tabular}} & \vcell{\begin{tabular}[b]{@{}c@{}}85.33\\\cellcolor{lightblue}(+0.53)\end{tabular}} & \vcell{} & \vcell{\begin{tabular}[b]{@{}c@{}}46.15\\\cellcolor{lightblue}(+2.19)\end{tabular}} & \vcell{\begin{tabular}[b]{@{}c@{}}86.63\\\cellcolor{lightblue}(+0.63)\end{tabular}} & \vcell{} & \vcell{\begin{tabular}[b]{@{}c@{}}36.02\\\cellcolor{lightblue}(+1.93)\end{tabular}} & \vcell{\begin{tabular}[b]{@{}c@{}}84.50\\\cellcolor{lightblue}(+0.71)\end{tabular}} \\[-\rowheight]
\printcelltop & \printcellmiddle & \printcellmiddle & \printcellmiddle & \printcellmiddle & \printcellmiddle & \printcellmiddle & \printcellmiddle & \printcellmiddle & \printcellmiddle & \printcellmiddle & \printcellmiddle & \printcellmiddle & \printcellmiddle & \printcellmiddle \\
\vcell{ALMA-13B-LoRA} & \vcell{\begin{tabular}[b]{@{}c@{}}27.19\\\cellcolor{lightblue}(+1.71)\end{tabular}} & \vcell{\begin{tabular}[b]{@{}c@{}}81.23\\\cellcolor{lightblue}(+1.02)\end{tabular}} & \vcell{} & \vcell{\begin{tabular}[b]{@{}c@{}}31.71\\\cellcolor{lightblue}(+0.45)\end{tabular}} & \vcell{\begin{tabular}[b]{@{}c@{}}84.68\\\cellcolor{lightblue}(+0.12)\end{tabular}} & \vcell{} & \vcell{\begin{tabular}[b]{@{}c@{}}40.00\\\cellcolor{lightred}(-0.26)\end{tabular}} & \vcell{\begin{tabular}[b]{@{}c@{}}85.43\\\cellcolor{lightblue}(+0.16)\end{tabular}} & \vcell{} & \vcell{\begin{tabular}[b]{@{}c@{}}46.45\\\cellcolor{lightblue}(+1.09)\end{tabular}} & \vcell{\begin{tabular}[b]{@{}c@{}}86.59\\\cellcolor{lightblue}(+0.12)\end{tabular}} & \vcell{} & \vcell{\begin{tabular}[b]{@{}c@{}}36.34\\\cellcolor{lightblue}(+0.75)\end{tabular}} & \vcell{\begin{tabular}[b]{@{}c@{}}84.48\\\cellcolor{lightblue}(+0.36)\end{tabular}} \\[-\rowheight]
\printcelltop & \printcellmiddle & \printcellmiddle & \printcellmiddle & \printcellmiddle & \printcellmiddle & \printcellmiddle & \printcellmiddle & \printcellmiddle & \printcellmiddle & \printcellmiddle & \printcellmiddle & \printcellmiddle & \printcellmiddle & \printcellmiddle \\
\vcell{text-davinci-003} & \vcell{\begin{tabular}[b]{@{}c@{}}27.10\\\cellcolor{lightblue}(+2.10)\end{tabular}} & \vcell{\begin{tabular}[b]{@{}c@{}}81.67\\\cellcolor{lightblue}(+0.05)\end{tabular}} & \vcell{} & \vcell{\begin{tabular}[b]{@{}c@{}}31.61\\\cellcolor{lightblue}(+0.73)\end{tabular}} & \vcell{\begin{tabular}[b]{@{}c@{}}84.67\\\cellcolor{lightred}(-0.12)\end{tabular}} & \vcell{} & \vcell{\begin{tabular}[b]{@{}c@{}}39.51\\\cellcolor{lightblue}(+1.04)\end{tabular}} & \vcell{\begin{tabular}[b]{@{}c@{}}85.52\\\cellcolor{lightblue}(+0.72)\end{tabular}} & \vcell{} & \vcell{\begin{tabular}[b]{@{}c@{}}46.71\\\cellcolor{lightblue}(+2.19)\end{tabular}} & \vcell{\begin{tabular}[b]{@{}c@{}}86.73\\\cellcolor{lightblue}(+0.57)\end{tabular}} & \vcell{} & \vcell{\begin{tabular}[b]{@{}c@{}}36.23\\\cellcolor{lightblue}(+1.52)\end{tabular}} & \vcell{\begin{tabular}[b]{@{}c@{}}84.65\\\cellcolor{lightblue}(+0.31)\end{tabular}} \\[-\rowheight]
\printcelltop & \printcellmiddle & \printcellmiddle & \printcellmiddle & \printcellmiddle & \printcellmiddle & \printcellmiddle & \printcellmiddle & \printcellmiddle & \printcellmiddle & \printcellmiddle & \printcellmiddle & \printcellmiddle & \printcellmiddle & \printcellmiddle \\
\vcell{GPT-4} & \vcell{\begin{tabular}[b]{@{}c@{}}27.20\\\cellcolor{lightblue}(+3.40)\end{tabular}} & \vcell{\begin{tabular}[b]{@{}c@{}}81.86\\\cellcolor{lightred}(-0.60)\end{tabular}} & \vcell{} & \vcell{\begin{tabular}[b]{@{}c@{}}32.71\\\cellcolor{lightblue}(+0.25)\end{tabular}} & \vcell{\begin{tabular}[b]{@{}c@{}}85.08\\\cellcolor{lightred}(-0.27)\end{tabular}} & \vcell{} & \vcell{\begin{tabular}[b]{@{}c@{}}42.17\\\cellcolor{lightblue}(+1.19)\end{tabular}} & \vcell{\begin{tabular}[b]{@{}c@{}}85.80\\\cellcolor{lightred}(-0.07)\end{tabular}} & \vcell{} & \vcell{\begin{tabular}[b]{@{}c@{}}49.83\\\cellcolor{lightblue}(+3.06)\end{tabular}} & \vcell{\begin{tabular}[b]{@{}c@{}}87.25\\\cellcolor{lightred}(-0.01)\end{tabular}} & \vcell{} & \vcell{\begin{tabular}[b]{@{}c@{}}37.73\\\cellcolor{lightblue}(+1.98)\end{tabular}} & \vcell{\begin{tabular}[b]{@{}c@{}}85.24\\\cellcolor{lightred}(-0.24)\end{tabular}} \\[-\rowheight]
\printcelltop & \printcellmiddle & \printcellmiddle & \printcellmiddle & \printcellmiddle & \printcellmiddle & \printcellmiddle & \printcellmiddle & \printcellmiddle & \printcellmiddle & \printcellmiddle & \printcellmiddle & \printcellmiddle & \printcellmiddle & \printcellmiddle \\
\bottomrule
\end{tabular} 
}
\caption{Performance of MT-Ladder on WMT22 XX$\rightarrow$En test set. The original translation using $\mathcal{P}_D$ prompt are at the top. The middle shows the MT-Ladder-2B refined scores, and the bottom shows the MT-Ladder-7B refined scores. \colorbox{lightblue}{Blue boxes} indicate improved MT-Ladder-refined scores, while \colorbox{lightred}{Red boxes} indicate decreased scores.
}
\label{tab:xxen}
\end{table*}

%% file: table/enxx_all.tex


\begin{table*}[ht]
\centering
\resizebox{\textwidth}{!}{%
\begin{tabular}{lcclcclcclcclcc} 
\toprule
\multirow{2}{*}{\textbf{Models}} & \multicolumn{2}{c}{\textbf{\textbf{En-Zh}}} & \multicolumn{1}{c}{} & \multicolumn{2}{c}{\textbf{En-De}} & \multicolumn{1}{c}{} & \multicolumn{2}{c}{\textbf{\textbf{\textbf{\textbf{En-Ru}}}}} &  & \multicolumn{2}{c}{\textbf{\textbf{\textbf{\textbf{En-Cs}}}}} & \multicolumn{1}{c}{} & \multicolumn{2}{c}{\textbf{Avg.}} \\ 
\cmidrule{2-3}\cmidrule{5-6}\cmidrule{8-9}\cmidrule{11-12}\cmidrule{14-15}
 & \textbf{BLEU} & \textbf{COMET} &  & \textbf{BLEU} & \textbf{COMET} &  & \textbf{BLEU} & \textbf{COMET} &  & \textbf{BLEU} & \textbf{COMET} &  & \textbf{BLEU} & \textbf{COMET} \\ 
\midrule
\multicolumn{15}{c}{\textit{Open}} \\
Alpaca-7B & 7.85 & 51.79 &  & 18.22 & 78.22 &  & 14.10 & 74.87 &  & 13.13 & 73.51 &  & 13.33 & 69.60 \\
Vicuna-7B-v1.5 & 31.42 & 82.68 &  & 22.65 & 80.82 &  & 19.60 & 81.07 &  & 16.37 & 77.25 &  & 22.51 & 80.46 \\
BayLing-13B & 37.93 & 84.63 &  & 25.62 & 82.70 &  & 12.77 & 71.01 &  & 16.43 & 78.22 &  & 23.19 & 79.14 \\
BigTranslate-13B & 29.89 & 81.83 &  & 22.99 & 80.54 &  & 19.52 & 81.56 &  & 22.68 & 84.50 &  & 23.77 & 82.11 \\
NLLB-3.3B & 32.53 & 81.57 &  & 33.97 & 86.24 &  & 30.11 & 87.51 &  & 36.30 & 89.90 &  & 33.23 & 86.31 \\
ALMA-7B-LoRA & 36.26 & 85.16 &  & 29.43 & 85.41 &  & 26.49 & 87.05 &  & 29.28 & 89.01 &  & 30.37 & 86.66 \\
ALMA-13B-LoRA & 39.87 & 85.96 &  & 31.49 & 85.62 &  & 29.03 & 87.53 &  & 32.47 & 89.79 &  & 33.22 & 87.23 \\ 
\hdashline
\multicolumn{15}{c}{\textit{Closed}} \\
text-davinci-003 & 38.34 & 85.76 &  & 31.85 & 85.61 &  & 27.55 & 86.74 &  & 31.28 & 88.57 &  & 32.26 & 86.67 \\
GPT-4 & 42.78 & 87.19 &  & 34.49 & 87.29 &  & 28.67 & 88.70 &  & 33.66 & 90.81 &  & 34.90 & 88.50 \\ 
\midrule
\multicolumn{15}{c}{\textit{MT-Ladder-2B Refinement}} \\
\vcell{Alpaca-7B} & \vcell{\begin{tabular}[b]{@{}c@{}}34.66\\\cellcolor{lightblue}(+26.81)\end{tabular}} & \vcell{\begin{tabular}[b]{@{}c@{}}83.56\\\cellcolor{lightblue}(+31.77)\end{tabular}} & \vcell{} & \vcell{\begin{tabular}[b]{@{}c@{}}24.81\\\cellcolor{lightblue}(+6.59)\end{tabular}} & \vcell{\begin{tabular}[b]{@{}c@{}}81.55\\\cellcolor{lightblue}(+3.33)\end{tabular}} & \vcell{} & \vcell{\begin{tabular}[b]{@{}c@{}}21.51\\\cellcolor{lightblue}(+7.41)\end{tabular}} & \vcell{\begin{tabular}[b]{@{}c@{}}83.71\\\cellcolor{lightblue}(+8.84)\end{tabular}} & \vcell{} & \vcell{\begin{tabular}[b]{@{}c@{}}20.62\\\cellcolor{lightblue}(+7.49)\end{tabular}} & \vcell{\begin{tabular}[b]{@{}c@{}}82.57\\\cellcolor{lightblue}(+9.06)\end{tabular}} & \vcell{} & \vcell{\begin{tabular}[b]{@{}c@{}}25.40\\\cellcolor{lightblue}(+12.07)\end{tabular}} & \vcell{\begin{tabular}[b]{@{}c@{}}82.85\\\cellcolor{lightblue}(+13.25)\end{tabular}} \\[-\rowheight]
\printcelltop & \printcellmiddle & \printcellmiddle & \printcellmiddle & \printcellmiddle & \printcellmiddle & \printcellmiddle & \printcellmiddle & \printcellmiddle & \printcellmiddle & \printcellmiddle & \printcellmiddle & \printcellmiddle & \printcellmiddle & \printcellmiddle \\
\vcell{Vicuna-7B-v1.5} & \vcell{\begin{tabular}[b]{@{}c@{}}36.47\\\cellcolor{lightblue}(+5.05)\end{tabular}} & \vcell{\begin{tabular}[b]{@{}c@{}}84.62\\\cellcolor{lightblue}(+1.94)\end{tabular}} & \vcell{} & \vcell{\begin{tabular}[b]{@{}c@{}}25.73\\\cellcolor{lightblue}(+3.08)\end{tabular}} & \vcell{\begin{tabular}[b]{@{}c@{}}81.86\\\cellcolor{lightblue}(+1.04)\end{tabular}} & \vcell{} & \vcell{\begin{tabular}[b]{@{}c@{}}22.59\\\cellcolor{lightblue}(+2.99)\end{tabular}} & \vcell{\begin{tabular}[b]{@{}c@{}}83.84\\\cellcolor{lightblue}(+2.77)\end{tabular}} & \vcell{} & \vcell{\begin{tabular}[b]{@{}c@{}}21.51\\\cellcolor{lightblue}(+5.14)\end{tabular}} & \vcell{\begin{tabular}[b]{@{}c@{}}83.19\\\cellcolor{lightblue}(+5.94)\end{tabular}} & \vcell{} & \vcell{\begin{tabular}[b]{@{}c@{}}26.58\\\cellcolor{lightblue}(+4.07)\end{tabular}} & \vcell{\begin{tabular}[b]{@{}c@{}}83.38\\\cellcolor{lightblue}(+2.92)\end{tabular}} \\[-\rowheight]
\printcelltop & \printcellmiddle & \printcellmiddle & \printcellmiddle & \printcellmiddle & \printcellmiddle & \printcellmiddle & \printcellmiddle & \printcellmiddle & \printcellmiddle & \printcellmiddle & \printcellmiddle & \printcellmiddle & \printcellmiddle & \printcellmiddle \\
\vcell{BayLing-13B} & \vcell{\begin{tabular}[b]{@{}c@{}}38.54\\\cellcolor{lightblue}(+0.61)\end{tabular}} & \vcell{\begin{tabular}[b]{@{}c@{}}85.03\\\cellcolor{lightblue}(+0.40)\end{tabular}} & \vcell{} & \vcell{\begin{tabular}[b]{@{}c@{}}26.71\\\cellcolor{lightblue}(+1.09)\end{tabular}} & \vcell{\begin{tabular}[b]{@{}c@{}}82.32\\\cellcolor{lightred}(-0.38)\end{tabular}} & \vcell{} & \vcell{\begin{tabular}[b]{@{}c@{}}21.67\\\cellcolor{lightblue}(+8.90)\end{tabular}} & \vcell{\begin{tabular}[b]{@{}c@{}}83.22\\\cellcolor{lightblue}(+12.21)\end{tabular}} & \vcell{} & \vcell{\begin{tabular}[b]{@{}c@{}}21.74\\\cellcolor{lightblue}(+5.31)\end{tabular}} & \vcell{\begin{tabular}[b]{@{}c@{}}82.93\\\cellcolor{lightblue}(+4.71)\end{tabular}} & \vcell{} & \vcell{\begin{tabular}[b]{@{}c@{}}27.17\\\cellcolor{lightblue}(+3.98)\end{tabular}} & \vcell{\begin{tabular}[b]{@{}c@{}}83.38\\\cellcolor{lightblue}(+4.24)\end{tabular}} \\[-\rowheight]
\printcelltop & \printcellmiddle & \printcellmiddle & \printcellmiddle & \printcellmiddle & \printcellmiddle & \printcellmiddle & \printcellmiddle & \printcellmiddle & \printcellmiddle & \printcellmiddle & \printcellmiddle & \printcellmiddle & \printcellmiddle & \printcellmiddle \\
\vcell{BigTranslate-13B} & \vcell{\begin{tabular}[b]{@{}c@{}}37.65\\\cellcolor{lightblue}(+7.76)\end{tabular}} & \vcell{\begin{tabular}[b]{@{}c@{}}84.74\\\cellcolor{lightblue}(+2.91)\end{tabular}} & \vcell{} & \vcell{\begin{tabular}[b]{@{}c@{}}26.82\\\cellcolor{lightblue}(+3.83)\end{tabular}} & \vcell{\begin{tabular}[b]{@{}c@{}}82.62\\\cellcolor{lightblue}(+2.08)\end{tabular}} & \vcell{} & \vcell{\begin{tabular}[b]{@{}c@{}}23.04\\\cellcolor{lightblue}(+3.52)\end{tabular}} & \vcell{\begin{tabular}[b]{@{}c@{}}84.03\\\cellcolor{lightblue}(+2.47)\end{tabular}} & \vcell{} & \vcell{\begin{tabular}[b]{@{}c@{}}24.39\\\cellcolor{lightblue}(+1.71)\end{tabular}} & \vcell{\begin{tabular}[b]{@{}c@{}}84.82\\\cellcolor{lightblue}(+0.32)\end{tabular}} & \vcell{} & \vcell{\begin{tabular}[b]{@{}c@{}}27.98\\\cellcolor{lightblue}(+4.21)\end{tabular}} & \vcell{\begin{tabular}[b]{@{}c@{}}84.05\\\cellcolor{lightblue}(+1.94)\end{tabular}} \\[-\rowheight]
\printcelltop & \printcellmiddle & \printcellmiddle & \printcellmiddle & \printcellmiddle & \printcellmiddle & \printcellmiddle & \printcellmiddle & \printcellmiddle & \printcellmiddle & \printcellmiddle & \printcellmiddle & \printcellmiddle & \printcellmiddle & \printcellmiddle \\
\vcell{NLLB-3.3B} & \vcell{\begin{tabular}[b]{@{}c@{}}39.06\\\cellcolor{lightblue}(+6.53)\end{tabular}} & \vcell{\begin{tabular}[b]{@{}c@{}}84.79\\\cellcolor{lightblue}(+3.22)\end{tabular}} & \vcell{} & \vcell{\begin{tabular}[b]{@{}c@{}}29.97\\\cellcolor{lightred}(-3.97)\end{tabular}} & \vcell{\begin{tabular}[b]{@{}c@{}}83.59\\\cellcolor{lightred}(-2.65)\end{tabular}} & \vcell{} & \vcell{\begin{tabular}[b]{@{}c@{}}25.03\\\cellcolor{lightred}(-5.08)\end{tabular}} & \vcell{\begin{tabular}[b]{@{}c@{}}85.19\\\cellcolor{lightred}(-2.32)\end{tabular}} & \vcell{} & \vcell{\begin{tabular}[b]{@{}c@{}}28.34\\\cellcolor{lightred}(-7.96)\end{tabular}} & \vcell{\begin{tabular}[b]{@{}c@{}}86.06\\\cellcolor{lightred}(-3.84)\end{tabular}} & \vcell{} & \vcell{\begin{tabular}[b]{@{}c@{}}30.60\\\cellcolor{lightred}(-2.63)\end{tabular}} & \vcell{\begin{tabular}[b]{@{}c@{}}84.91\\\cellcolor{lightred}(-1.40)\end{tabular}} \\[-\rowheight]
\printcelltop & \printcellmiddle & \printcellmiddle & \printcellmiddle & \printcellmiddle & \printcellmiddle & \printcellmiddle & \printcellmiddle & \printcellmiddle & \printcellmiddle & \printcellmiddle & \printcellmiddle & \printcellmiddle & \printcellmiddle & \printcellmiddle \\ 
\midrule
\multicolumn{15}{c}{\textit{MT-Ladder-7B Refinement}} \\
\vcell{BigTranslate-13B} & \vcell{\begin{tabular}[b]{@{}c@{}}42.10\\\cellcolor{lightblue}(+12.21)\end{tabular}} & \vcell{\begin{tabular}[b]{@{}c@{}}86.56\\\cellcolor{lightblue}(+4.73)\end{tabular}} & \vcell{} & \vcell{\begin{tabular}[b]{@{}c@{}}32.00\\\cellcolor{lightblue}(+9.01)\end{tabular}} & \vcell{\begin{tabular}[b]{@{}c@{}}85.92\\\cellcolor{lightblue}(+5.38)\end{tabular}} & \vcell{} & \vcell{\begin{tabular}[b]{@{}c@{}}28.11\\\cellcolor{lightblue}(+8.59)\end{tabular}} & \vcell{\begin{tabular}[b]{@{}c@{}}87.38\\\cellcolor{lightblue}(+5.82)\end{tabular}} & \vcell{} & \vcell{\begin{tabular}[b]{@{}c@{}}30.49\\\cellcolor{lightblue}(+7.81)\end{tabular}} & \vcell{\begin{tabular}[b]{@{}c@{}}89.00\\\cellcolor{lightblue}(+4.50)\end{tabular}} & \vcell{} & \vcell{\begin{tabular}[b]{@{}c@{}}33.18\\\cellcolor{lightblue}(+9.41)\end{tabular}} & \vcell{\begin{tabular}[b]{@{}c@{}}87.22\\\cellcolor{lightblue}(+5.11)\end{tabular}} \\[-\rowheight]
\printcelltop & \printcellmiddle & \printcellmiddle & \printcellmiddle & \printcellmiddle & \printcellmiddle & \printcellmiddle & \printcellmiddle & \printcellmiddle & \printcellmiddle & \printcellmiddle & \printcellmiddle & \printcellmiddle & \printcellmiddle & \printcellmiddle \\
\vcell{NLLB-3.3B} & \vcell{\begin{tabular}[b]{@{}c@{}}43.40\\\cellcolor{lightblue}(+10.87)\end{tabular}} & \vcell{\begin{tabular}[b]{@{}c@{}}86.65\\\cellcolor{lightblue}(+5.08)\end{tabular}} & \vcell{} & \vcell{\begin{tabular}[b]{@{}c@{}}33.33\\\cellcolor{lightred}(-0.64)\end{tabular}} & \vcell{\begin{tabular}[b]{@{}c@{}}86.34\\\cellcolor{lightblue}(+0.10)\end{tabular}} & \vcell{} & \vcell{\begin{tabular}[b]{@{}c@{}}29.55\\\cellcolor{lightred}(-0.56)\end{tabular}} & \vcell{\begin{tabular}[b]{@{}c@{}}87.71\\\cellcolor{lightblue}(+0.20)\end{tabular}} & \vcell{} & \vcell{\begin{tabular}[b]{@{}c@{}}33.74\\\cellcolor{lightred}(-2.56)\end{tabular}} & \vcell{\begin{tabular}[b]{@{}c@{}}89.37\\\cellcolor{lightred}(-0.53)\end{tabular}} & \vcell{} & \vcell{\begin{tabular}[b]{@{}c@{}}35.01\\\cellcolor{lightblue}(+1.78)\end{tabular}} & \vcell{\begin{tabular}[b]{@{}c@{}}87.52\\\cellcolor{lightblue}(+1.21)\end{tabular}} \\[-\rowheight]
\printcelltop & \printcellmiddle & \printcellmiddle & \printcellmiddle & \printcellmiddle & \printcellmiddle & \printcellmiddle & \printcellmiddle & \printcellmiddle & \printcellmiddle & \printcellmiddle & \printcellmiddle & \printcellmiddle & \printcellmiddle & \printcellmiddle \\
\vcell{ALMA-7B-LoRA} & \vcell{\begin{tabular}[b]{@{}c@{}}42.17\\\cellcolor{lightblue}(+5.91)\end{tabular}} & \vcell{\begin{tabular}[b]{@{}c@{}}86.73\\\cellcolor{lightblue}(+1.57)\end{tabular}} & \vcell{} & \vcell{\begin{tabular}[b]{@{}c@{}}32.33\\\cellcolor{lightblue}(+2.90)\end{tabular}} & \vcell{\begin{tabular}[b]{@{}c@{}}86.20\\\cellcolor{lightblue}(+0.79)\end{tabular}} & \vcell{} & \vcell{\begin{tabular}[b]{@{}c@{}}28.58\\\cellcolor{lightblue}(+2.09)\end{tabular}} & \vcell{\begin{tabular}[b]{@{}c@{}}87.65\\\cellcolor{lightblue}(+0.60)\end{tabular}} & \vcell{} & \vcell{\begin{tabular}[b]{@{}c@{}}30.90\\\cellcolor{lightblue}(+1.62)\end{tabular}} & \vcell{\begin{tabular}[b]{@{}c@{}}89.30\\\cellcolor{lightblue}(+0.29)\end{tabular}} & \vcell{} & \vcell{\begin{tabular}[b]{@{}c@{}}33.50\\\cellcolor{lightblue}(+3.13)\end{tabular}} & \vcell{\begin{tabular}[b]{@{}c@{}}87.47\\\cellcolor{lightblue}(+0.81)\end{tabular}} \\[-\rowheight]
\printcelltop & \printcellmiddle & \printcellmiddle & \printcellmiddle & \printcellmiddle & \printcellmiddle & \printcellmiddle & \printcellmiddle & \printcellmiddle & \printcellmiddle & \printcellmiddle & \printcellmiddle & \printcellmiddle & \printcellmiddle & \printcellmiddle \\
\vcell{ALMA-13B-LoRA} & \vcell{\begin{tabular}[b]{@{}c@{}}42.72\\\cellcolor{lightblue}(+2.85)\end{tabular}} & \vcell{\begin{tabular}[b]{@{}c@{}}86.83\\\cellcolor{lightblue}(+0.87)\end{tabular}} & \vcell{} & \vcell{\begin{tabular}[b]{@{}c@{}}32.54\\\cellcolor{lightblue}(+1.05)\end{tabular}} & \vcell{\begin{tabular}[b]{@{}c@{}}85.93\\\cellcolor{lightblue}(+0.31)\end{tabular}} & \vcell{} & \vcell{\begin{tabular}[b]{@{}c@{}}29.04\\\cellcolor{lightblue}(+0.01)\end{tabular}} & \vcell{\begin{tabular}[b]{@{}c@{}}87.65\\\cellcolor{lightblue}(+0.12)\end{tabular}} & \vcell{} & \vcell{\begin{tabular}[b]{@{}c@{}}31.70\\\cellcolor{lightred}(-0.77)\end{tabular}} & \vcell{\begin{tabular}[b]{@{}c@{}}89.43\\\cellcolor{lightred}(-0.36)\end{tabular}} & \vcell{} & \vcell{\begin{tabular}[b]{@{}c@{}}34.00\\\cellcolor{lightblue}(+0.79)\end{tabular}} & \vcell{\begin{tabular}[b]{@{}c@{}}87.46\\\cellcolor{lightblue}(+0.24)\end{tabular}} \\[-\rowheight]
\printcelltop & \printcellmiddle & \printcellmiddle & \printcellmiddle & \printcellmiddle & \printcellmiddle & \printcellmiddle & \printcellmiddle & \printcellmiddle & \printcellmiddle & \printcellmiddle & \printcellmiddle & \printcellmiddle & \printcellmiddle & \printcellmiddle \\
\vcell{text-davinci-003} & \vcell{\begin{tabular}[b]{@{}c@{}}43.62\\\cellcolor{lightblue}(+5.28)\end{tabular}} & \vcell{\begin{tabular}[b]{@{}c@{}}86.75\\\cellcolor{lightblue}(+0.99)\end{tabular}} & \vcell{} & \vcell{\begin{tabular}[b]{@{}c@{}}32.90\\\cellcolor{lightblue}(+1.05)\end{tabular}} & \vcell{\begin{tabular}[b]{@{}c@{}}86.12\\\cellcolor{lightblue}(+0.51)\end{tabular}} & \vcell{} & \vcell{\begin{tabular}[b]{@{}c@{}}28.58\\\cellcolor{lightblue}(+1.03)\end{tabular}} & \vcell{\begin{tabular}[b]{@{}c@{}}87.92\\\cellcolor{lightblue}(+1.18)\end{tabular}} & \vcell{} & \vcell{\begin{tabular}[b]{@{}c@{}}32.57\\\cellcolor{lightblue}(+1.29)\end{tabular}} & \vcell{\begin{tabular}[b]{@{}c@{}}89.25\\\cellcolor{lightblue}(+0.68)\end{tabular}} & \vcell{} & \vcell{\begin{tabular}[b]{@{}c@{}}34.42\\\cellcolor{lightblue}(+2.16)\end{tabular}} & \vcell{\begin{tabular}[b]{@{}c@{}}87.51\\\cellcolor{lightblue}(+0.84)\end{tabular}} \\[-\rowheight]
\printcelltop & \printcellmiddle & \printcellmiddle & \printcellmiddle & \printcellmiddle & \printcellmiddle & \printcellmiddle & \printcellmiddle & \printcellmiddle & \printcellmiddle & \printcellmiddle & \printcellmiddle & \printcellmiddle & \printcellmiddle & \printcellmiddle \\
\vcell{GPT-4} & \vcell{\begin{tabular}[b]{@{}c@{}}44.35\\\cellcolor{lightblue}(+1.57)\end{tabular}} & \vcell{\begin{tabular}[b]{@{}c@{}}87.02\\\cellcolor{lightred}(-0.17)\end{tabular}} & \vcell{} & \vcell{\begin{tabular}[b]{@{}c@{}}33.81\\\cellcolor{lightred}(-0.68)\end{tabular}} & \vcell{\begin{tabular}[b]{@{}c@{}}86.55\\\cellcolor{lightred}(-0.74)\end{tabular}} & \vcell{} & \vcell{\begin{tabular}[b]{@{}c@{}}29.32\\\cellcolor{lightblue}(+0.65)\end{tabular}} & \vcell{\begin{tabular}[b]{@{}c@{}}88.15\\\cellcolor{lightred}(-0.55)\end{tabular}} & \vcell{} & \vcell{\begin{tabular}[b]{@{}c@{}}32.65\\\cellcolor{lightred}(-1.01)\end{tabular}} & \vcell{\begin{tabular}[b]{@{}c@{}}89.69\\\cellcolor{lightred}(-1.12)\end{tabular}} & \vcell{} & \vcell{\begin{tabular}[b]{@{}c@{}}35.03\\\cellcolor{lightblue}(+0.13)\end{tabular}} & \vcell{\begin{tabular}[b]{@{}c@{}}87.85\\\cellcolor{lightred}(-0.65)\end{tabular}} \\[-\rowheight]
\printcelltop & \printcellmiddle & \printcellmiddle & \printcellmiddle & \printcellmiddle & \printcellmiddle & \printcellmiddle & \printcellmiddle & \printcellmiddle & \printcellmiddle & \printcellmiddle & \printcellmiddle & \printcellmiddle & \printcellmiddle & \printcellmiddle \\
\bottomrule
\end{tabular} 
}
\caption{Results of MT-Ladder on WMT22 En$\rightarrow$XX test set. MT-Ladder-2B refines LLMs with higher parameter counts than itself. MT-Ladder-7B refines all translators except for GPT-4. The marker are the same in Table~\ref{tab:xxen}. 
}

\label{tab:enxx}
\end{table*}

%% file: table/refine_cmp_v2.tex

\begin{table}
\centering
\resizebox{\columnwidth}{!}{%
\begin{tabular}{lccccccc} 
\toprule
\multicolumn{1}{l}{\multirow{2}{*}{\textbf{Models}}} & \multicolumn{7}{c}{\textbf{\textbf{\textbf{\textbf{\textbf{\textbf{\textbf{\textbf{COMET}}}}}}}}}                                                                                                                                                                                                                                                                                                   \\ 
\cmidrule{2-8}
\multicolumn{1}{l}{}                                 & \textbf{\textbf{\textbf{\textbf{\textbf{\textbf{\textbf{\textbf{Zh-En}}}}}}}} & \multicolumn{1}{l}{} & \textbf{\textbf{\textbf{\textbf{En-Zh}}}} & \multicolumn{1}{l}{} & \textbf{\textbf{\textbf{\textbf{De-En}}}} & \multicolumn{1}{l}{} & \textbf{\textbf{\textbf{\textbf{\textbf{\textbf{\textbf{\textbf{\textbf{\textbf{\textbf{\textbf{\textbf{\textbf{\textbf{\textbf{En-De}}}}}}}}}}}}}}}}  \\ 
\midrule
\multicolumn{1}{l}{Palm2}                            & 74.70                                                                         & \multicolumn{1}{l}{} & -                                         & \multicolumn{1}{l}{} & -                                         & \multicolumn{1}{l}{} & 81.80                                                                                                                                                  \\
\multicolumn{1}{r}{+LLMRefine}                                           & 75.90                                                                         & \multicolumn{1}{l}{} & -                                         & \multicolumn{1}{l}{} & -                                         & \multicolumn{1}{l}{} & 82.30                                                                                                                                                  \\ 
\midrule
\multicolumn{1}{l}{BigTranslate-13B}                 & 74.63                                                                         &                      & 81.83                                     &                      & 81.04                                     &                      & 80.54                                                                                                                                                  \\
\multicolumn{1}{r}{+Vicuna-13B-v1.5 (Re)}                                & 72.53                                                                         &                      & 80.91                                     &                      & 77.26                                     &                      & 78.79                                                                                                                                                  \\
\multicolumn{1}{r}{+Vicuna-13B-v1.5 (CT)}                                & 76.53                                                                         &                      & 83.67                                     &                      & 81.72                                     &                      & 81.05                                                                                                                                                  \\
\multicolumn{1}{r}{+TowerInstruct-7B}                                    & 76.17                                                                         &                      & 85.62                                     &                      & 82.03                                     &                      & 84.89                                                                                                                                                  \\
\multicolumn{1}{r}{+TowerInstruct-13B}                                   & 77.92                                                                         &                      & 85.91                                     &                      & 82.26                                     &                      & 85.86                                                                                                                                                  \\
\multicolumn{1}{r}{+MT-Ladder-2B}                                        & 79.28                                                                         &                      & 84.74                                     &                      & 83.45                                     &                      & 82.62                                                                                                                                                  \\
\multicolumn{1}{r}{+MT-Ladder-7B}                                        & \uline{81.08}                                                                 &                      & \uline{86.56}                             &                      & \uline{84.58}                             &                      & \uline{85.92}                                                                                                                                          \\
\multicolumn{1}{r}{+GPT-4o mini}                                         & \textbf{81.34}                                                                &                      & \textbf{86.57}                            &                      & \textbf{84.70}                            &                      & \textbf{86.30}                                                                                                                                         \\
\bottomrule
\end{tabular}
}
\caption{Comparison with baselines on WMT22 test set. Palm2 and LLMRefine results are from \citet{xu2023pinpoint}. Contrast Translation (CT) and Rephrase (Re) are two prompt-based strategies from \citet{chen2023iterative}. \textbf{Bold font} and \uline{underline} indicate the best and second best performance, respectively. }
\label{tab:refine_cmp_v2}
\end{table}

%% file: table/change_base.tex

\begin{table}[ht]
\centering
\resizebox{\columnwidth}{!}{%
\begin{tabular}{ccclcc} 
\toprule
\multicolumn{3}{c}{\textbf{MT-Ladder Pipeline }}                                           &  & \multicolumn{2}{c}{\textbf{WMT22 En-Zh}}             \\ 
\cmidrule{1-3}\cmidrule{5-6}
\textbf{Sampling Model}                 & \textbf{Base Model}  & \textbf{Refine Model} &  & \textbf{BLEU} & \textbf{COMET} \\ 
\midrule
\multirow{2}{*}{Gemma-2B-it}    & \multirow{2}{*}{Gemma-2B}   & Gemma-2B-it        &  & 35.46         & 84.41                     \\
                                &                                  & Gemma-7B-it        &  & 35.86         & 84.60                   \\ 
\hdashline
\multirow{2}{*}{Vicuna-7B-v1.5} & \multirow{2}{*}{LLaMA-2-7B} & Vicuna-7B-v1.5     &  & 34.31         & 84.12                 \\
                                &                                  & Vicuna-13B-v1.5    &  & 36.19         & 84.74                \\ 
\midrule 
\midrule
\multicolumn{6}{c}{\textit{Baseline }}   
\\
\midrule 
\midrule
\multicolumn{4}{l}{Gemma-2B-it}                                                            & 21.07         & 78.67               \\
\multicolumn{4}{l}{Gemma-7B-it}                                                            & 30.55         & 81.50                  \\
\multicolumn{4}{l}{Vicuna-7B-v1.5}                                                         & 31.42         & 82.68               \\
\multicolumn{4}{l}{Vicuna-13B-v1.5}                                                        & 35.14         & 83.38           \\
\bottomrule
\end{tabular}
}
\caption{Ablation of different sampling and backbones. Evaluate Gemma and LLaMA suite models on En-Zh.}
\vspace{-3mm}
\label{tab:change_base}
\end{table}

%% file: appendix/data_parallel.tex
\begin{table*}[ht]
\centering
\resizebox{0.7\textwidth}{!}{%
\begin{tabular}{lcccc} 
\toprule
\multirow{2}{*}{Language} & \multicolumn{4}{c}{Parallel Data}                              \\ 
\cmidrule{2-5}
                          & Train & Development & Test (from English) & Test (to English)  \\ 
\midrule
Chinese (Zh)              & 15406 & 1002        & 2037                & 1875               \\
German (De)                & 14211 & 1002        & 2037                & 1984               \\
Russia (Ru)               & 15000 & 1002        & 2037                & 2016               \\
Czech (Cs)                & 12076 & 1002        & 2037                & 1448               \\
\bottomrule
\end{tabular}
}
\caption{The statistics for the parallel data we used.}
\label{tab:data}
\end{table*}

%% file: appendix/hft_stage_case.tex
\begin{table*}[ht]
\centering
\resizebox{\textwidth}{!}{%
\begin{tabular}{|>{\raggedright\arraybackslash}m{0.3\textwidth}|>{\raggedright\arraybackslash}m{0.5\textwidth}|>{\raggedright\arraybackslash}m{0.2\textwidth}|}
\hline
\multicolumn{2}{|c|}{\textbf{Anti-HFT Case}} & \textbf{COMET} \\ \hline
\textbf{German Source} & So jedenfalls macht die grandiose F1-Saison wesentlich weniger Spaß als es mit einem vernünftigen Sender möglich wäre. & - \\ \hline
\textbf{English Reference} & At any rate, it really makes the grand F1 season considerably less fun as would be the case with a reasonable broadcaster. & 95.61 \\ \hline
\textbf{Intermediate Translation} & So, in any case, the grandiose F1 season is much less fun than it would be with a reasonable broadcaster. & 87.50 \\ \hline
\textbf{Anti-HFT Stage1 \space \space \space \space \space \space \space \space \space \space \space \space (Hard)} & So, at any rate, the grandiose F1 season is much less fun than it would be with a reasonable broadcaster. & 87.55 \\ \hline
\textbf{Anti-HFT Stage2 (Hard+Medium)} & So, at least, the grandiose F1 season is much less fun than it would be with a reasonable broadcaster. & 83.32 \\ \hline
\textbf{Anti-HFT Stage3 (Hard+Medium+Easy)} & So the great F1 season is much less fun than it would be with a decent broadcaster. & 81.57 \\ \hline
\multicolumn{2}{|c|}{\textbf{HFT Cases}} & \textbf{COMET} \\ \hline
\textbf{German Source} & Es ist schade, dass wir den Flow nicht mitnehmen konnten. & - \\ \hline
\textbf{English Reference} & It is a shame that we were not able to get into the flow. & 96.32 \\ \hline
\textbf{Intermediate Translation} & It is a shame that we couldn't take the flow with us. & 83.21 \\ \hline
\textbf{HFT Stage1 \space \space \space \space \space \space \space \space \space \space \space \space \space \space \space \space \space \space \space \space \space \space \space \space (Easy)} & It's a shame we couldn't keep the momentum going. & 79.54 \\ \hline
\textbf{HFT Stage2 \space \space \space \space \space \space (Easy+Medium)} & It's a shame that we couldn't take the flow with us. & 81.18 \\ \hline
\textbf{HFT Stage3 (Easy+Medium+Hard)} & It's a shame that we couldn't keep the flow going. & 84.10 \\ \hline
\end{tabular}%
}
\caption{Case study. Stage corresponds to Figure \ref{fig:anti_trend}.}
\label{tab:stage_case}
\end{table*}

%% file: appendix/hierarchical_case.tex
\begin{CJK}{UTF8}{gkai}
\begin{table*}[ht]
\centering
\resizebox{\textwidth}{!}{%
\begin{tabular}{ll}
\hline
\multicolumn{2}{c}{\textbf{COMET:69.73}}                                                                                           \\ \hline
\textbf{Chinese Source}           & 但八年前濒临倒闭，不得不接受救助从那时开始便放弃了那样的追求。                                                                \\
\textbf{Intermediate Translation} &
  \begin{tabular}[c]{@{}l@{}}But eight years ago, it was on the verge of bankruptcy and had to accept help.\\ From that time on, I gave up such pursuits.\end{tabular} \\
\textbf{English Reference}                & It has retreated from them since it nearly collapsed eight years ago and had to be bailed out. \\ \hline
\multicolumn{2}{c}{\textbf{COMET:83.37}}                                                                                           \\ \hline
\textbf{English Source} &
  \begin{tabular}[c]{@{}l@{}}Representatives of junior doctors have called on their union to authorise fresh industrial action \\ in their dispute about a new contract.\end{tabular} \\
\textbf{Intermediate Translation} & 低级医生代表呼吁他们的工会授权新的工业行动，因为他们对新合同的争议仍未得到解决。                                                       \\
\textbf{Chinese Reference}                & 初级医生代表号召联盟批准其针对新合同纠纷采取新的劳工行动。                                                                  \\ \hline
\multicolumn{2}{c}{\textbf{COMET:91.84}}                                                                                           \\ \hline
\textbf{German Source}            & Ich hätte mich gefreut, wenn Mesut Özil weiter für Deutschland gespielt hätte.                 \\
\textbf{Intermediate Translation} & I would have been delighted if Mesut Özil had continued to play for Germany.                   \\
\textbf{English Reference}                & I would be happy if Mesut Özil continued to play for Germany.                                  \\ \hline
\end{tabular}%
}
\caption{Cases of triples with different COMET scores.}
\label{tab:hierarchy_case}
\end{table*}
\end{CJK}